\pdfoutput=1

\documentclass[11pt]{article}

\usepackage{acl}

\usepackage{times}
\usepackage{latexsym}

\usepackage[T1]{fontenc}

\usepackage[utf8]{inputenc}

\usepackage{microtype}
\usepackage{enumitem}
\usepackage{colortbl}
\usepackage{xcolor}
\usepackage{esvect}
\usepackage{url}
\usepackage{booktabs}
\usepackage{scrextend}
\usepackage{graphicx}
\usepackage{textpos}
\usepackage[ruled]{algorithm}
\usepackage[noend]{algpseudocode}
\usepackage[normalem]{ulem}
\usepackage{siunitx}
\usepackage{amsmath}
\usepackage{amssymb} 
\usepackage{multirow}
\usepackage{lscape}
\usepackage{rotating}
\usepackage{xspace}
\usepackage{wasysym} 
\usepackage[capitalise]{cleveref}
\usepackage{tablefootnote}
\usepackage{pifont}
\usepackage{phaistos}
\usepackage{soul}
\usepackage{subcaption}
\usepackage{makecell}
\usepackage{listofitems}
\usepackage{transparent}
\usepackage{readarray}
\usepackage{environ}
\usepackage{tabstackengine}
\usepackage{etoolbox}
\usepackage{pgfplots}
\pgfplotsset{compat=1.18}
\usepackage{tikz}
\usepackage{tikzscale}
\usetikzlibrary{decorations.pathmorphing,decorations.pathreplacing,calligraphy}
\usetikzlibrary{shapes.callouts}

\newcommand{\cmark}{\ding{56}\xspace}%
\newcommand{\ymark}{\ding{51}}
\newcommand{\xmark}{\ding{55}}

\makeatletter
\newcommand{\rAB}[2]{\bgroup \UL@setULdepth
 \markoverwith{\lower\ULdepth\hbox
   {\kern-.03em\vbox{\color{#1}\hrule width.2em\kern1.2\p@\color{#2}\hrule}\kern-.03em}}%
 \ULon}
\makeatother
\definecolor{cyelllow}{HTML}{FFC300}
\definecolor{ablue}{HTML}{006795}
\definecolor{jgreen}{HTML}{98CC70}
\definecolor{xgbblue}{HTML}{006795}
\definecolor{dbertagreen}{HTML}{3a6a00}
\definecolor{cblue}{HTML}{4d4dff}

\newcommand{\ctest}[0]{C-Test\xspace}
\newcommand{\ctests}[0]{C-Tests\xspace}

\title{Constrained C-Test Generation via Mixed-Integer Programming}

\author{
    Ji-Ung Lee$^{1,\dag}$ \quad
    Marc E. Pfetsch$^{2}$ \quad 
    Iryna Gurevych$^{1,\dag}$
    \\
    $^{1}$Ubiquitous Knowledge Processing Lab, Department of Computer Science \\
    $^{2}$ Research Group Optimization, Department of Mathematics \\ 
    Technical University of Darmstadt, $^{\dag}$Hessian AI 
}

\begin{document}
\maketitle
\begin{abstract}
This work proposes a novel method to generate \ctests; a deviated form of cloze tests (a gap filling exercise) where only the last part of a word is turned into a gap.
In contrast to previous works that only consider varying the gap size or gap placement to achieve locally optimal solutions, we propose a mixed-integer programming (MIP) approach.
This allows us to consider gap size and placement simultaneously, achieving globally optimal solutions, and to directly integrate state-of-the-art models for gap difficulty prediction into the optimization problem.
A user study with 40 participants across four \ctest generation strategies (including GPT-4) shows that our approach (\texttt{MIP}) significantly outperforms two of the baseline strategies (based on gap placement and GPT-4); and performs on-par with the third (based on gap size). 
Our analysis shows that GPT-4 still struggles to fulfill explicit constraints during generation and that \texttt{MIP} produces \ctests that correlate best with the perceived difficulty. 
We publish our code, model, and collected data consisting of 32 English \ctests with 20 gaps each (totaling 3,200 individual gap responses) under an open source license.\footnote{\url{https://github.com/UKPLab/arxiv2024-constrained-ctest-generation}}
\end{abstract}

\section{Introduction}\label{sec:introduction}

Learning a (second) language is one of the key factors that can directly impact a person's success in life.
It enables them to participate in daily and social life and can even grant them new job opportunities. 
A crucial part of language learning is expanding the vocabulary by learning new words and their correct inflection. 
Gap filling exercises provide one possibility to consolidate new words and practice grammar rules~\citep{oller1973cloze}.
Whereas most works focus on cloze tests~\citep{Taylor1953}, a gap filling exercise where a whole word is turned into a gap, we focus on \ctests, a deviated cloze test~\citep{Klein1982}.
In contrast to cloze tests, \ctests only remove the last part of a word, leaving the rest as a hint (cf.\ \cref{fig:constraints-example}).
This reduces the ambiguity of gap filling compared to cloze tests and requires learners to actively inflect words themselves. 
To provide sufficient context, the first and the last sentences remain free of gaps.

\begin{figure*}[t!]
    \begin{small}
    \include{images/mip_vs_neural}
    \end{small}
    \vspace{-2em} 
    \caption{A simplified \ctest generation example. Colors indicate the gap sizes and words considered during generation. While \texttt{SIZE} ({$\color{red}\blacksquare$}) only varies the gap size with a static placement (every second word) and \texttt{SEL} ({$\color{blue}\blacksquare$}) only the placement with a static gap size (the second half of a word, rounded up), \texttt{MIP} ({$\color{blue!50!red}\transparent{0.5}\blacksquare$}) considers all possible combinations. In contrast to \texttt{MIP}, purely neural approaches (\texttt{GPT-4}) provide no theoretical guarantee that all constraints are always satisfied. In this example, the word \textit{on} is fully turned into a gap although the model correctly states in its response that words are only ``partially deleted'' in \ctests (cf.\,\cref{fig:gpt-4-simple-example} for the full prompt and response).}
  \label{fig:constraints-example}
\end{figure*}

Despite above advantages, a major limitation of \ctests is the prevalent static generation strategy that turns every second half of every second word into a gap.
This impairs the usability of \ctests for two reasons.
First, the difficulty of a \ctest solely depends on the underlying text.
Second, to suit a learner's curriculum, \ctests need to be adapted post-generation; for instance, by manually placing gaps at words that consider a newly learned vocabulary. 
While novel \ctest generation strategies are necessary to tackle these issues, a key challenge is the large number of possible \ctests that can be generated from a single text.
For instance, placing $m$ gaps among $n$ words already results in $\binom{n}{m}=\frac{n!}{m! (n-m)!}$ possible \ctests (cf. \cref{sec:appendix-brute-forcing}).
To reduce the complexity of the task, \citet{lee-etal-2019-manipulating} manually restrict the number of possible \ctests and only vary the gap size with a static gap placement (or vice versa).
Although they successfully generate \ctests with varying difficulties using the same text, this discards a large number of \ctests as potential solutions.
Moreover, their strategies iteratively select the best gap size or placement which does not take the global interdependencies between gaps into account, leading to locally optimal solutions.

We instead propose to tackle \ctest generation as a mixed-integer programming (MIP) problem which results in three advantages over existing methods.\footnote{We provide a primer on MIP in \cref{sec:appendix-mip-primer}.}
First, we can now make use of well-established solving methods that efficiently remove whole sets of unsuitable \ctests while finding a provably optimal solution (see, e.g., \citealt{schrijver1986theory}).
This allows us to consider \textit{all} possible \ctests (instead of only a fraction) within a feasible run time of $\sim$48.6 seconds.\footnote{In \cref{sec:appendix-objective}, we devise methods that further reduce the run time to $\sim$3.1 seconds.}
Second, we can integrate trained gap difficulty prediction models into the optimization problem and solve the whole problem in an end-to-end manner~\citep{bunel-2018-mip,anderson2020strong}.
Third, in contrast to (purely neural) large language models (LLMs), the use of MIP provides a theoretical guarantee that the resulting \ctest always satisfies all constraints such as the number of gaps or their size (cf. \cref{fig:constraints-example}).
Together, these advantages allow teachers to directly generate \ctests that suit their needs; eliminating the need to adapt them post-generation.
Our contributions are:\looseness=-1
\begin{itemize}
    \item A novel generation method for \ctests (\texttt{MIP}), that combines state-of-the-art models for gap difficulty prediction with constrained optimization methods.
    \item A user study with 40 participants that compares \texttt{MIP} against the gap size and placement generation strategies, as well as against \ctests generated by GPT-4.
    \item Our data consisting of 32 \ctests with 20 gaps each, their respective error-rates, as well as perceived difficulties on a exercise-level.
\end{itemize}

\section{Related Work}\label{sec:rel_work}
Various works have shown the usefulness of \ctests in second language learning scenarios~\citep{Chapelle94,BABAII2001209,grotjahn-2006,mckay2019more}.
To select \ctests that suit a learner's curriculum, it is necessary to predict their difficulty; i.e., proficient learners should receive more difficult \ctests than inexperienced learners. 
This is done by either directly predicting difficulty of the whole \ctest~\citep{10.1162/tacl_a_00310,mccarthy-etal-2021-jump}, or---for a more fine-grained selection---by aggregating the individual gap difficulties~\citep{lee-etal-2020-empowering}.
Predicting the difficulty of individual gaps is thus key, with past works investigating a wide range of features such as word frequency or readability scores across different models~\citep{brown1989cloze,Sigott1995,eckes2011item,beinborn-2014-ctest,Beinborn2016}.

Instead of selecting suited \ctests, one could also directly generate \ctests with a specific difficulty.
Towards this end, \citet{lee-etal-2019-manipulating} propose automated generation strategies for \ctests based on a single text.
Although they only vary the gap size or placement at once due to the large number of possible \ctests, another advantage of deviating from the static generation strategy is a better quantification of a learner's proficiency~\citep{clearly-1988-xtest,Kamimoto1993KJ00007033409,laufer-1999-ctest}.

Finally, recent advances in LLMs have led to the emergence of new learning opportunities for students and teachers~\citep{kohnke-2023-chatgpt}.
Despite mixed results in essay scoring~\citep{naismith-etal-2023-automated,yancey-etal-2023-rating} and feedback generation~\citep{duenas-etal-2023-youve,wang-demszky-2023-chatgpt}, LLMs are easy to access and to use which makes them a tempting alternative to proprietary and expensive educational resources. 
As such, \citet{xiao-etal-2023-evaluating} investigate ChatGPT~\citep{instructGPT-2022} to generate reading comprehension exercises and find that the model struggles to generate appropriate distractors (incorrect answers in multiple-choice questions).
As \ctests are solely based on deletion, generating them may be an easier task for LLMs than exercises that also require the generation of distractors. 
We thus include GPT-4~\citep{openai2023gpt4} as a baseline in our user study, but find that the model struggles to fulfill hard constraints such as the number of gaps.

In contrast, our method uses MIP---a more general form of integer linear programming (ILP)---which allows us to find globally optimal solutions (instead of locally optimal ones, cf.\,\citealt{lee-etal-2019-manipulating}) by relying upon constrained optimization methods which have been successfully applied across various NLP tasks~\citep{roth-yih-2004-linear,barzilay-lapata-2006-aggregation,martins-2009-ilp,koo-etal-2010-dual,berant-etal-2011-global,lin-ng-2021-system}.

\section{MIP Definition}\label{sec:prelim}
Our goal is to generate a \ctest from a text $\mathcal{T}$ with a target difficulty $\tau$ and $m$ gaps $g$. 
The gaps are selected from a set $\mathcal{G} \subset \mathcal{T}$ that denotes all words which can be turned into a gap (e.g., excluding first and last sentences).
We further define $\tau \in [0, 1]$ to be the error-rate computed over the whole \ctest:
\begin{equation*}
\tau = \frac{1}{m} \sum^{m}_{i=1} \texttt{error}(g_i), 
\end{equation*}
where $g_i$ denotes the $i$-th gap of the \ctest. 
The function $\texttt{error}(\cdot)$ indicates if the $i$-th gap was filled-out correctly and returns a binary value (0 for correct and 1 for incorrect). 
Consequently, smaller values of $\tau$ relate to easier, and larger values to more difficult \ctests. 
As the actual $\texttt{error}(\cdot)$ function is learner dependent and not known during generation, we approximate it using a gap difficulty prediction model $f_{\theta}: \mathbb{R}^{k} \mapsto [0, 1]$ with parameters $\theta$ that computes the error-rate for each gap, represented as a $k$-dimensional, real number vector~$\textbf{x}$.\footnote{From here on, we will refer to the gap error-rate as the error-rate and the \ctest error-rate as the difficulty for clarity.} 
We can now define the estimated difficulty $\hat{\tau}$ for any selection of $m$ gaps $g \in \mathcal{G}$:
\begin{equation*}
\hat{\tau} = \frac{1}{m} \sum^{m}_{i=1} f_{\theta}(g_i).
\end{equation*}
Given $f_{\theta}$, any \ctest that has minimal distance between the estimated and the target difficulties is optimal.
Hence, our optimization objective is:
\begin{eqnarray*}\label{eq:basic} 
    \min & |\tau - \hat{\tau}|.
\end{eqnarray*}

\paragraph{Gap placement.} 
So far, $\hat{\tau}$ only includes gaps that have already been selected.
To model the task of optimally placing them across all possible gaps $\mathcal{G}$ with $|\mathcal{G}|=n > m$, we now introduce binary decision variables: 
\begin{flalign*}
    \min_{ b_i \in \{0,1\}} \quad &  |\tau - \frac{1}{m} \sum^{n}_{i=1} b_{i}\,  f_{\theta}(g_i)|  \\
    \textrm{s.t.} \quad & \sum^{n}_{i=1}{b_i} = m, 
\end{flalign*}
where $b_i$ denotes a binary decision variable for a selected gap at the $i$-th word.\footnote{Note, that $i$ now ranges from 1 to $n$, with $n$ denoting the number of all possible gaps and $m$ the target number of gaps.}  
The constraint ensures that the resulting \ctest has exactly $m$ gaps. 

\paragraph{Gap size.}
In addition to the placement, each gap can assume a different size.
We hence extend our objective with additional binary decision variables $s_{i,j}$ for the gap size where $l_i$ denotes the length of the $i$-th word with $j \in \{1,\dots,l_i - 1\}$. 
Our final model comprising gap placement and gap size is then:
\begin{flalign}
    \min_{ s_{i,j},b_i \in \{0,1\}} \quad &  |\tau - \frac{1}{m} \sum^{n}_{i=1} b_{i}  \sum^{l_i -1}_{j=1} s_{i,j}f_{\theta}(g_{i,j})|  \label{eq:objective} \\
    \textrm{s.t.} \quad & \sum^{n}_{i=1}{b_i} = m, \label{eq:constraint-placement} \\
     & \sum^{l_i-1}_{j=1}{s_{i,j}} = 1, \label{eq:constraint-size} 
\end{flalign}
\noindent
where $g_{i,j}$ denotes a gap at the $i$-th word with size $j$, for all words $i \in \{1,\dots, n\}$.
Our binary decision variables $b_i$ and $s_{i,j}$ for the gap placement and size are constrained by \cref{eq:constraint-placement} and \cref{eq:constraint-size}, respectively. 

\paragraph{Considerations.}
Analyzing our final model reveals three traits about the \ctest generation task and how we have defined it. 
First, we see that the number of possible gap placements is already very large with $\binom{n}{m}$; making it intractable to try out all combinations to identify an optimal \ctest with certainty. 
Second, despite the large number of possible combinations, the search space is still finite due to the discrete values of gap size and placement.
Consequently, it is possible that two \ctests are equally optimal with the same estimated difficulty $\hat{\tau}$.
Third, we formulate the objective in a general way which does not include any learner-specific terms.
Although this work does not investigate a learner-specific adaptation, we note that the generation process can be tailored to a specific group of learners with a model trained on learner-specific data, or by adding learner-specific constraints (e.g., one that limits the gap size to a specific value). 

\section{Task Instantiation}\label{sec:approach}
Having defined our general optimization objective, we need to identify a suitable model $f_{\theta}$ for our evaluation study and define its constraints accordingly. 
As we consider $f_{\theta}$ including all its parameters in our objective, it needs to be small enough so that the whole optimization problem remains solvable in feasible time.
We thus focus on the feature-based model proposed by \citet{Beinborn2016} for predicting the gap error-rate which achieves state-of-the-art performance.
In addition to the 59 features utilized by \citet{Beinborn2016}, we further include two additional BERT-based features that have been shown to be helpful~\citep{lee-etal-2020-empowering}.
A single gap $g_{i,j}$ is thus represented by a $61$-dimensional vector $\textbf{x}$, where each dimension $k$ relates to a specific feature.
As most of the features (51) do not depend on the gap size $s_{i,j}$ or placement $b_i$, we can model them as equality constraints for our model $f_{\theta}$:
\begin{flalign}
     & g_{i,j,k} = c_{i,j,k}, \label{eq:constraint-constants} 
\end{flalign}
\noindent
where $c_{i,j,k}$ is the $k$-th pre-computed feature value for gap $g_{i,j,k}$.\footnote{We provide a detailed description of all features in \cref{sec:appendix-feature-description} and an ablation study in \cref{sec:appendix-ablation}.}

\subsection{Gap Size Features}

Overall, we identify six features that change depending on the gap size $s_{i,j}$. 
The first two are the BERT-based features that estimate the certainty that BERT correctly predicts the masked gap~\citep{lee-etal-2020-empowering}.
For this, the authors use the probability of BERT predicting the correct solution ($k=49$) and the entropy of the normalized vector of prediction probabilities for the top-50 candidates ($k=50$).
Next, we have three binary features that measure whether the gap occurs at a compound break, i.e., if the gap and the non-gap part are words on their own ($k=56$), whether the non-gap part only consists of \textit{th} ($k=57$), and whether the gap begins at a syllable break ($k=58$).
Finally, we adapt the \textit{WordLengthInCharacters}~(\citealt{Beinborn2016}, page 220) to model the varying gap size ($k=59$), providing our model with a proper notion of gap size.
We now describe our features that depend on the gap size in relation to our gap size decision variable $s_{i,j}$ by adding the following constraint to the MIP:
\begin{flalign}
     & g_{i,j,k} = \textbf{s}_i \cdot \text{C}_{i,k}, \label{eq:constraint-gap-size} 
\end{flalign}
\noindent
where $\textbf{s}_i \in \mathbb{B}^{l_i-1}$ is the 1-hot vector (of length $l_i-1$) representing the gap size decision variables (with 1 being at the $j$-th position) for gap $g_i$.
The matrix $\text{C}_{i,k} \in \mathbb{R}^{l_i-1 \times k}$ represents all possible values $k \in \mathcal{U}$ can take for all possible gap sizes $j \in \{1,\dots,l_i-1\}$ at gap $g_i$ with $\mathcal{U} = \{49, 50, 56, 57, 58, 59\}$; i.e., all our features that depend on the gap size.

\subsection{Gap Placement Features}

We identify four features that change depending on the gap placement. 
The first feature indicates if the word occurs somewhere else in the \ctest as a gap ($k=51$). 
The second one represents the number of gaps in the same sentence ($k=52$).
The third and fourth features measure the number of preceding gaps in the \ctest ($k=53$) and in the same sentence ($k=54$).  
In contrast to the gap size features, all placement features depend on the placement of the other gaps.
We thus need to model these dependencies into our constraints:
\begin{flalign}
 & g_{i,j,51} = \max(\textbf{b}\cdot\text{V}), \label{eq:constraint-placement-1} \\  
 & g_{i,j,52} = \sum_{h \in \mathcal{S}_i} b_h, \label{eq:constraint-placement-2}  \\  
 & g_{i,j,53} = \sum_{h=1}^{i-1} b_{h}, \label{eq:constraint-placement-3} \\   
  & g_{i,j,54} = \sum_{h \in \mathcal{S}_i, h < i} b_h,  \label{eq:constraint-placement-4}   
\end{flalign}
for all $i,j \in \{1,\dots,n\}$ where $\mathcal{S}_i$ denotes the set of all words in the sentence containing $g_i$.
The vector \textbf{b} denotes all placement decision variables $b_i$ and V the $n\times n$ matrix of binary values $v_{i,j}$ with:
\[
    v_{i,j} = 
\begin{cases}
    1, & \text{if } w_i = w_j, \\
    0, & \text{otherwise},
\end{cases} 
\]
for all $i,j \in \{1,\dots,n\}$ where $i \neq j$.

\section{Gap Difficulty Model}\label{sec:experiments}
With \cref{eq:objective}--\eqref{eq:constraint-placement-4} defining our full optimization model, we focus on training a well-performing regression model $f_{\theta}$ that predicts the gap error-rate.
To ensure that the optimization model remains solvable in feasible time, we focus on small models with architectures that provide strong guarantees~\citep{anderson2020strong}.

\begin{table}[!t]
    \centering
    \begin{small}
    \begin{tabular}{llrrrr}
        \toprule
        Dataset & Usage & \# CT & \# Gaps & $\mu$ GS & $\sigma$ GS \\ 
        \midrule 
        ACL$_{\text{2020-train}}$ & Train & 69 & 1,480 & 2.80 & 1.30 \\
        ACL$_{\text{2020-test}}$ & Dev & 5 & 100 & 2.95 & 1.50 \\
        ACL$_{2019}$ & Test & 16 & 320 & 2.56 & 1.76 \\
        \bottomrule
    \end{tabular}
    \end{small}
    \caption{Dataset statistics. \# CT is the number of \ctests per dataset. $\mu$ GS is the average gap size and $\sigma$ GS the respective standard deviation. Each \ctest has 20 gaps.}
    \label{tab:ctest-data-statistics}
\end{table}

\paragraph{Data.} We use two datasets in total:
\begin{description}[noitemsep,topsep=3pt,itemsep=3pt,itemindent=-1em]
\item[ACL-2020] \citet{lee-etal-2020-empowering} provided us with their dataset for training our models.\footnote{We received permission to use the data for training and sharing our models (but not the data). We contacted authors of other works but without success due to proprietary restrictions.} 
It consists of 69 \ctests we use for training, and 5 \ctests (their test set) which we use as our development set. 
The data was collected by their university's language learning center from students taking language assessment tests and consists of gap error-rates for \ctests generated with the static generation strategy.
\item[ACL-2019] \citet{lee-etal-2019-manipulating} published 16 \ctests collected in their user study under a creative commons license. 
This is the only available data which contains gap error-rates for \ctests deviating from the static generation strategy, generated by \texttt{SEL} and \texttt{SIZE} (explained in the next section). We use this dataset as the test set to identify the best model for varying gap sizes and placements.
\end{description}

\cref{tab:ctest-data-statistics} shows the number of \ctests, gaps (i.e., instances), and the data splits.
We can see that the ACL-2019 data differs substantially from the ACL-2020 data as it has the smallest gap sizes on average ($\mu$ GS), but the largest standard deviation ($\sigma$ GS).\looseness=-1

\paragraph{Experimental setup.}
We consider four different model types for $f_{\theta}$.
Linear regression models (LR), support vector machines with a linear kernel (SVM), multi-layer perceptrons (MLP), and gradient boosted trees (XGB).
We further include more recent models in our evaluation, namely, the base and large versions of BERT~\citep{devlin-etal-2019-bert}, RoBERTa~\citep{liu2019roberta}, and DeBERTa~\citep{he2021deberta}.
Each of the LR, SVM, MLP, and XGB models are trained to predict the gap error-rate given the 61 respective features as the input.
We evaluate three setups for the transformer-based models, namely, a masked-regression (MR) setup, a CLS-token setup, and a feature-enriched CLS-token setup (\cref{sec:appendix-transformers}).
Following \citet{beinborn-2014-ctest}, we use the root mean squared error (RMSE) and Pearson's correlation $\rho$ to evaluate all models. 
We use an \textit{Intel Core™ i5-8400 CPU} with 6 x 2.80GHz for training the LR, SVM, MLP, and XGB models and a single NVIDIA A100 with 80\,GB for training the transformer-based models.
We provide details on the training parameters and about hyperparameter tuning in \cref{sec:appendix-hyperparameter}.

\begin{table}[!t]
    \centering
    \begin{small}
    \begin{tabular}{lrrrr}
        \toprule
        Model&  $\downarrow$ RMSE & $\uparrow$ Pearson's $\rho$ \\
        \midrule
        SVM$_{\text{Linear}, c=0.01}$ &  1.846 & $-0.111$ \\
        MLP$_{\text{ReLU}}$ & 1.113 & 0.099 \\
        MLP$_{\text{Linear}}$ &  1.681 & 0.085 \\
        LR &  2.949 & $-0.385$ \\
        XGB & \textbf{0.285} & \textbf{0.439} \\ 
        \midrule
        BERT$_{\text{base}}$  & 0.311 & 0.279 \\
        BERT$_{\text{large}}$  & 0.319 & 0.174 \\
        RoBERTa$_{\text{base}}$  & 0.324 & 0.159 \\
        RoBERTa$_{\text{large}}$  & 0.324 & 0.050 \\
        DeBERTa$_{\text{v3-base}}$  & 0.311 & 0.245  \\
        DeBERTa$_{\text{v3-large}}$  & 0.308 & 0.259 \\
        \bottomrule
    \end{tabular}
    \end{small}
    \caption{Root mean squared error (RMSE) and Pearson's correlation $\rho$ for predicting the gap error-rate on our test data. Overall, we find that XGB performs best, even outperforming large masked language models.} 
    \label{tab:results-model-tuning}
\end{table}

\paragraph{Results.}
\cref{tab:results-model-tuning} shows the results (averaged across ten runs) on the test data for the best performing models and transformer setup based on our hyperparameter tuning.
Overall, we find that XGB performs best and is the only feature-based model that handles the varying gap sizes and placements well, as all other feature-based models show substantially worse scores.
An analysis reveals that they may have overfitted on the training data, as they perform substantially better on the development data that closer matches the training data in terms of average gap size and variance.
We further find a similar tendency for the feature-enriched transformer models (\cref{tab:results-model-tuning-full}).
We conjecture that XGB may be more robust to these changes due to its tree structure and the ensembling performed during training.
Finally, we find a robust performance for the transformer-based models trained using only the CLS-token or in an MR manner; with a worse performance than the feature-based models on the development data, but a higher performance on the test data.
Conducting an error analysis on the model with the best RMSE on the test data (DeBERTa$_{\text{v3-large}}$) reveals that the model performs better than XGB for small gap sizes (one and three), but worse for gap sizes equal or larger than four.
We provide the full results and a detailed error analysis in \cref{sec:appendix-full-results-intrinsic}.
We select the best performing model (XGB) for our user study.

\section{User Study}\label{sec:study}
To evaluate our optimization model, we conduct a user study where we ask our participants to complete \ctests. 
Our main goal is to evaluate the following research hypothesis: \\

\noindent
\textit{\ctests that have been generated with our approach have a smaller distance} $|\tau - \tau^{*}|$ \textit{(i.e., between the target difficulty} $\tau$ \textit{and the observed difficulty} $\tau^{*}$ \textit{) than \ctests that have been generated with other approaches.}
\\

For a single \ctest, the observed difficulty is defined as:
\begin{equation}
    \tau^{*} = \frac{1}{m \cdot r} \sum^{m}_{i=1}\sum^{r}_{j=1} \texttt{error}(g_{i,j}),
\end{equation}
where $r$ is the number of individual responses for the $i$-th gap and $m$ the total number of gaps in the \ctest. 
As before, $\texttt{error}(g_{i,j})$ returns 1 if $g_{i,j}$ was filled out incorrectly and 0 otherwise.
We focus on the observed difficulty $\tau^{*}$, as our goal is to evaluate the performance of the generation strategies and not the difficulty prediction model, which would require a different study setup (i.e., a study centered around quantifying $|\hat{\tau} - \tau^{*}|$).

\subsection{Setup}
Following \citet{lee-etal-2019-manipulating}, we generate \ctests with $m=20$ gaps out of $n=40$ possible words.
We select two target difficulties $\tau \in \{0.1, 0.9\}$ as target difficulties that have been shown to be either easy ($\tau = 0.1$) or difficult ($\tau = 0.9$) for the baseline strategies to achieve~\citep{lee-etal-2019-manipulating}.

\paragraph{Generation strategies.} We compare our approach (\texttt{MIP}) against three generation strategies based on gap placement (\texttt{SEL}), size (\texttt{SIZE}) \citep{lee-etal-2019-manipulating} and \texttt{GPT-4}~\citep{openai2023gpt4}.\footnote{To select strong baselines, we  reimplemented \texttt{SEL} and \texttt{SIZE} using the XGB model but found that the original models performed better. We provide results and detailed descriptions of all models in \cref{sec:appendix-reimplementation}.}

\begin{description}[noitemsep,topsep=3pt,itemsep=3pt,itemindent=-1.5em]
\item[\texttt{SEL}] The gap placement strategy first estimates all gap error-rates for all $g \in \mathcal{G}$ in text $\mathcal{T}$ where the second half of a word is turned into a gap.
It then iteratively selects the gaps with the smallest distance to $\tau$; alternating between gaps that are easier and harder than $\tau$. 
\item[\texttt{SIZE}] The gap size strategy places gaps according to the static generation strategy. 
For each gap, the respective error-rate is then computed and a character is removed or added iteratively until the target difficulty $\tau$ is reached. 
To increase computational efficiency, two separate models are trained on a reduced set of features which predict the relative change in difficulty for increasing or decreasing the gap size.
\item[\texttt{GPT-4}] Finally, we use GPT-4 to generate \ctests. For a fair comparison, we only show the text passage containing $\mathcal{G}$ to the model (hence, no gaps are placed in the first and last sentence) and use five instances from the ACL-2019 dataset as few-shot examples in our prompt. We provide details on the few-shot example selection, \ctest generation, and the full final prompt in \cref{sec:appendix-gpt-example}.\footnote{Note, that we cannot use the other datasets for GPT-4 due to the signed non-disclosure agreements regarding the data.} 
\end{description}

For \texttt{MIP}, we use Gurobi as a popular off-the-shelf solver (\citealt{gurobi}) to generate the \ctests (with an average run time of 48.6 seconds). 
We provide a fine-grained analysis with respect to run time in \cref{sec:appendix-runtime-mip} and devise further improvements (reducing the average run time to 3.1 seconds) using different formulations of the optimization objective in \cref{sec:appendix-objective}.

\paragraph{Text selection.}
We select text passages from four public domain books indexed at project Gutenberg.\footnote{\url{https://www.gutenberg.org/}}
Considering that all of our participants have an academic background, we select books with a high reading difficulty (\cref{tab:reading-difficulties}).
From each book, we randomly sample passages that contain at least three sentences and at least 40 words that can be turned into a gap (e.g., words that have at least two characters, do not contain any numbers, etc). 
We further avoid passages that contain dialogues. 
Data pre-processing (sentence splitting and tokenization) is done using NLTK's \texttt{sent\_tokenize} and \texttt{word\_tokenize} functions~\citep{bird2009natural}. 

\subsection{Design}  
Constructing the study with four texts $\mathcal{T}_1$--$\mathcal{T}_4$, two target difficulties $\tau \in \{0.1, 0.9\}$, and four generation strategies results in 32 \ctests (eight \ctests for each generation strategy and text).
To prevent participants from memorizing parts of the solution, each participant can only receive \ctests generated from different texts, i.e., each participant is asked to solve four \ctests. 

\paragraph{Groups.}
To decide upon the groups (consisting of four \ctests) to which our participants are assigned, we need to consider the respective configuration (i.e., which strategies and difficulties the group is composed of).
Even a fixed ordering of the texts amounts to too many configurations to cover all---especially, as we also require multiple measurements for each \ctest to account for random effects such as our participants' individual performance.
To obtain stable estimates that allow us to compare different generation strategies against each other, we construct our groups following a Latin Hypercube Experimental Design principle~\citep{latinhypercube-1979}.
This ensures that we select configurations with a minimal overlap between all possible combinations of text, strategy, and difficulty---with each text and strategy occurring once, and each target difficulty twice in all configurations (details in \cref{sec:appendix-hypercube}).

\paragraph{Procedure.}
We implement and host a study interface using Flask\footnote{\url{https://flask.palletsprojects.com/}} and SQLAlchemy\footnote{\url{https://www.sqlalchemy.org/}}.
On the landing page, participants are asked to enter a self-chosen study key of which we store the respective hash (so that we cannot guess a participant from the key). 
The key is only used in case a participant decides to access, change, or delete their data after the study.
Upon registration, participants are informed about the purpose of the study, the collected data, and its use and are asked for their consent to participate in the study.
Participants are further asked five questions about their English proficiency and are shown an example \ctest before being randomly assigned to one of the eight groups.
After each of the four \ctests, we ask our participants to provide a self-assessment of the perceived difficulty of the \ctest on a 5-point Likert scale and to estimate the number of gaps filled-out correctly.
\cref{fig:questionnaire,fig:study-ctest,fig:study-feedback} show the questionnaire, the study interface, and the collected feedback to each \ctest, respectively.

\subsection{Results}
Overall, we recruited 40 volunteers for our study, resulting in five responses for each of the 32 \ctests.
All participants have at least B1 proficiency on the CEFR scale~\citep{cefr} with a majority having C1 (16) or C2 (13) proficiency.
Most of our participants are native German speakers (26) with the remainder distributed across 11 other languages.\footnote{We provide detailed statistics in \cref{sec:appendix-study-participants}.} 
On average, participants spent $\sim$4.5 minutes per \ctest and $\sim$17 minutes to finish all four \ctests.
\cref{tab:ctest-results-coarse} shows the average distance ($\mu$) between the target difficulty $\tau$ and observed difficulty $\tau^{*}$ as well as the standard deviation ($\sigma$) and variance ($\sigma^2$).\footnote{We compute the observed difficulty as the \ctest error-rate, i.e., the fraction of incorrect gaps over all 20 gaps.}
While we can observe substantial differences between different generation strategies, we also see high standard deviations (larger than 0.2), making it difficult to interpret the results.

\begin{table}[!t]
    \centering
    \begin{tabular}{lll} 
    \toprule
        $|\tau - \tau^{*}|$ &  $\mu$  & $\sigma$ \\ 
        \midrule
        \texttt{GPT-4} & 0.45$^*$ & 0.23 \\ 
        \texttt{MIP} & 0.36 & 0.29 \\ 
        \texttt{SEL} & 0.39$^*$ & 0.27 \\ 
        \texttt{SIZE} & \textbf{0.34}$^\circ$ & 0.29 \\ 
         \bottomrule
    \end{tabular}\caption{Average ($\mu$) and standard deviation ($\sigma$) of $|\tau - \tau^{*}|$  for all generation strategies (lower is better). 
    The Wald test~\citep{wald1943tests} shows that \texttt{MIP} performs significantly better than \texttt{GPT-4} and \texttt{SEL} ($^*$), but find no significant differences to \texttt{SIZE} ($^\circ$).}\label{tab:ctest-results-coarse}
\end{table}

\paragraph{Statistical significance.}
To test if the observed differences with respect to the generation strategies are statistically significant, we fit a generalized additive mixed model (GAMM, \citealt{lin1999inference}) on our collected data. 
This allows us to concurrently model our response variable ($\tau^{*}$) on a continuous latent scale using multiple predictor variables expressed as a sum of smooth functions of covariates while accounting for random effects. 
Using the GAMM fitted on our data, we can now test for statistical significance between different \ctest generation strategies using the Wald test~\citep{wald1943tests}.
Overall, we find that all strategies significantly outperform \texttt{GPT-4} which has the highest $|\tau - \tau^{*}|$ (cf.~\cref{tab:ctest-results-coarse}).
In addition, \texttt{MIP} and \texttt{SIZE} significantly outperform \texttt{SEL}.
Although \texttt{MIP} performs slightly worse than \texttt{SIZE}, we find that the differences are not statistically significant.
Interestingly, we find that \ctests generated using \texttt{GPT-4} show a reverse tendency for $\tau^{*}$, displaying a significantly higher difficulty for $\tau=0.1$ than for $\tau=0.9$ ($p < 0.001$).
In other words, our participants made significantly more errors on \texttt{GPT-4}-generated \ctests that were supposed to be easier and vice versa.\footnote{Detailed statistics and the formular of the GAMM are provided in \cref{sec:appendix-study-gamm}.}

\begin{table}[!htb]
    \centering
    \begin{small}
    \begin{tabular}{lcccc}
    \toprule
        $\tau$ &  \texttt{GPT-4} & \texttt{MIP} & \texttt{SEL} & \texttt{SIZE} \\ 
        \midrule
        0.1 & 4.3$\pm0.56$ & \textbf{2.6}$\pm0.97$ & 3.5$\pm0.74$ & 3.0$\pm1.16$ \\ 
        0.9 & 4.0$\pm0.56$ & 3.9$\pm0.70$ & \textbf{4.2}$\pm0.60$ & 3.8$\pm0.83$ \\ 
        \bottomrule
    \end{tabular}
    \end{small}
    \caption{Average perceived difficulty on a 5-point Likert scale between too easy (1) and too hard (5). }\label{tab:ctest-feedback-coarse} %
\end{table}

\paragraph{Perceived difficulty.}
\cref{tab:ctest-feedback-coarse} shows the average estimates of the perceived \ctest difficulty on a 5-point Likert scale~\citep{likert1932technique} between \textit{too easy} (1) and \textit{too hard} (5). 
Overall, we find the largest difference in perceived difficulty for \texttt{MIP} (1.3; i.e., more than one rating), indicating that \texttt{MIP} generated \ctests that were perceived substantially easier (or harder) for $\tau=0.1$ ($\tau=0.9$).  
Whereas we find similar tendencies for \texttt{SEL} and \texttt{SIZE}, \ctests generated with \texttt{GPT-4} again show a reversed tendency for the perceived difficulty (i.e., our participants found presumably easy \ctests harder than the difficult ones).
To check for statistical significance, we fit a second GAMM for ordinal data and find that \texttt{MIP} is again significantly better than \texttt{SEL} and \texttt{GPT-4} with no significant difference to \texttt{SIZE}.\footnote{We provide details on the second GAMM and box plots for the perceived difficulties in \cref{sec:appendix-study-gamm-feedback}.}

\begin{figure*}[!htb]
  \begingroup
  \small
  \arrayrulecolor{gray}
  \newcolumntype{L}[1]{>{\ttfamily\small\raggedright\arraybackslash}m{#1}}
  \begin{tabular}{|*{2}{@{ }L{.49\textwidth}@{ }|}}
    \hline
    \dots As little\_{$\color{jgreen}\blacksquare$} as the act of birth comes into consideration in the who\_\_{$\color{jgreen}\blacksquare$} proces\_{$\color{jgreen}\blacksquare$} a\_\_{$\color{jgreen!90!red}\blacksquare$} procedur\_{$\color{jgreen}\blacksquare$} of heredit\_{$\color{jgreen!90!red}\blacksquare$}, ju\_\_{$\color{jgreen}\blacksquare$} as littl\_{$\color{jgreen}\blacksquare$} is ``being-conscious'' oppose\_{$\color{jgreen!90!red}\blacksquare$} t\_{$\color{jgreen}\blacksquare$} the instinctiv\_{$\color{jgreen}\blacksquare$} in a\_\_{$\color{jgreen!50!red}\blacksquare$} d\_\_\_\_\_\_\_{$\color{red}\blacksquare$} sen\_\_{$\color{jgreen!80!red}\blacksquare$}; the grea\_\_\_{$\color{jgreen!90!red}\blacksquare$} p\_\_\_{$\color{jgreen}\blacksquare$} of the consciou\_{$\color{jgreen!90!red}\blacksquare$} thi\_\_\_\_\_{$\color{jgreen!90!red}\blacksquare$} of a philosophe\_{$\color{jgreen!90!red}\blacksquare$} i\_{$\color{jgreen!80!red}\blacksquare$} secretly influenced by his instincts\dots
    &
    \dots As lit\_\_\_{$\color{jgreen}\blacksquare$} as t\_\_{$\color{jgreen}\blacksquare$} act o\_{$\color{jgreen}\blacksquare$} birth com\_\_{$\color{jgreen}\blacksquare$} into consider\_\_\_\_\_{$\color{jgreen}\blacksquare$} in t\_\_{$\color{jgreen}\blacksquare$} whole pro\_\_\_\_{$\color{jgreen}\blacksquare$} and proc\_\_\_\_\_{$\color{jgreen!60!red}\blacksquare$} of here\_\_\_\_{$\color{jgreen!90!red}\blacksquare$}, just a\_{$\color{jgreen}\blacksquare$} little i\_{$\color{jgreen!90!red}\blacksquare$} ``being-conscious'' opposed to t\_\_{$\color{jgreen}\blacksquare$} instinctive i\_{$\color{jgreen!90!red}\blacksquare$} any deci\_\_\_\_{$\color{jgreen!60!red}\blacksquare$} sense; t\_\_{$\color{jgreen}\blacksquare$} greater pa\_\_{$\color{jgreen}\blacksquare$} of t\_\_{$\color{jgreen}\blacksquare$} conscious thin\_\_\_\_{$\color{jgreen}\blacksquare$} of a philo\_\_\_\_\_\_{$\color{jgreen}\blacksquare$} is secr\_\_\_\_{$\color{jgreen!90!red}\blacksquare$} influneced by his instincts\dots
    \\ \hline
    \multicolumn{1}{c}{(a) C-test of $\mathcal{T}_1$ generated with \texttt{MIP} for $\tau = 0.1$} & 
    \multicolumn{1}{c}{(b) C-test of $\mathcal{T}_1$ generated with \texttt{SIZE} for $\tau = 0.1$}
    \\
  \end{tabular}
  \endgroup
  \caption{\texttt{MIP} vs \texttt{SIZE}. Colored squares indicate the gap error-rates (0.0 {$\color{jgreen}\blacksquare$}
{$\color{jgreen!90!red}\blacksquare$}
{$\color{jgreen!80!red}\blacksquare$}
{$\color{jgreen!60!red}\blacksquare$}
{$\color{jgreen!50!red}\blacksquare$}
{$\color{red}\blacksquare$} 1.0)}
  \label{fig:mip-vs-size}
\end{figure*}

\subsection{Error Analysis}

\paragraph{\texttt{MIP} vs. \texttt{SIZE}.}
Analyzing the \ctests generated by \texttt{MIP} reveals that the generation strategy may struggle with assessing interdependencies between gaps.
For instance, we find that the \ctest generated from $\mathcal{T}_1$ with $\tau=0.1$ contains multiple successive gaps (\cref{fig:mip-vs-size}, left).
This substantially increases the difficulty compared to the \texttt{SIZE} strategy that only places gaps at every second word (\cref{fig:mip-vs-size}, right).
Intuitively, gaps that occur in succession should be harder to fill out---e.g., consider a \ctest where a single sentence only contains gaps versus a \ctest where the gaps are evenly distributed.
However, we find no patterns with respect to the length and occurrences of successive gaps for \ctests with varying difficulties (\cref{tab:mip-vs-size-chains}).
We identify three causes for this shortcoming and discuss potential solutions to be addressed in future work.
Each cause can either be attributed to the difficulty prediction model $f_{\theta}$ (i.e., the XGB model) or to the optimization model (i.e., \texttt{MIP}).
First, $f_{\theta}$ has only been trained on \ctests generated by the static strategy which leads to a lack of successive gaps during training; despite the decent performance on the test data with different gap sizes and placements.
This work alleviates this issue by providing data with varying gap sizes and placements for training. 
Second, $f_{\theta}$ only uses features that implicitly capture the interdependency such as the number of gaps in a sentence.
This could be tackled by explicitly modeling interdependencies; e.g., with a binary feature that indicates if the previous word is a gap.
Finally, \texttt{MIP} does not specifically model interdependencies. 
One way to better capture interdependencies could be to introduce a weighting term in our objective (\cref{eq:objective}) and increase the estimated gap error-rate according to the number of successive gaps. 

\begin{table}[!ht]
    \centering
    \begin{small}
    \begin{tabular}{llllll}
    \toprule
       & $\tau$ & $\mathcal{T}_1$& $\mathcal{T}_2$& $\mathcal{T}_3$ & $\mathcal{T}_4$ \\ 
        \midrule
      \# Gaps & 0.1 & 20$^5$ & 20 & 20$^2$ & 20$^2$ \\ 
       & 0.9 & 29 & 25 & 37 & 22 \\ 
         \midrule
      $\mu$ Size & 0.1 &  3.4 & 4.1 & 3.6 & 3.7 \\ 
       & 0.9 & 3.45 & 3.75 & 2.9 & 3.4 \\ 
         \bottomrule
    \end{tabular}
    \end{small}
    \caption{Number of gaps and average size for \ctests generated by \texttt{GPT-4}. Superscripts denote the number of required regenerations to obtain at least 20 gaps. }\label{tab:gpt-4-errors}
\end{table}

\paragraph{Shortcomings of \texttt{GPT-4}.}
The increasing use of GPT-4 (and ChatGPT) in education was a key reason to include it in the study~\citep{zhang2023small}, making it even more concerning that the model performed worst.
Analyzing the \ctests generated by \texttt{GPT-4} reveals that gaps are frequently clustered at the beginning for $\tau=0.9$.
\cref{tab:gpt-4-errors} shows the number of generated gaps per text (\# Gaps) and their average size ($\mu$ Size) after pruning to 20 gaps. 
We find that the model generates substantially more gaps for $\tau=0.9$, which shows that it lacks a notion of gap-level difficulty and simply adds more gaps to increase the difficulty.
Moreover, the model generates convincing (but incorrect) explanations along with the exercise (cf. \cref{sec:appendix-error-analysis}), which could be especially harmful in self-learning scenarios with only GPT-4 as the tutor.
This highlights the importance of approaches that better control the output of LLMs~\citep{zhang2023-controlled-generation}.

\paragraph{Tuning the prompt.} 
To better understand the extent of above issues, we explored more sophisticated prompting strategies (after the study) with the goal to improve GPT-4's notion of \ctest difficulty. 
First, we asked the model to generate multiple \ctests at once with both target difficulties.
Second, we provided corrective feedback while asking it to increase or decrease the difficulty.
First and foremost, we find that GPT-4 has undergone substantial changes in the meantime, as it provides very different responses.
Most notably, the model now prioritizes a modification of the gap size to increase or decrease the \ctest difficulty.
Despite this improvement, the key issues remain as there are still instances where whole words are turned into gaps or where the model decreases the gap size when prompted to increase the difficulty. 
Interestingly, we find that GPT-3.5 provides responses that are similar to those during study development; with a model that mostly aims to control the difficulty with the number of placed gaps.
Overall, we conclude that both models still struggle to follow hard constraints and moreover, that they lack an inherent notion of \ctest difficulty.

\section{Conclusion}\label{sec:conclusion}
This work proposes a first constrained generation strategy for \ctests, a type of gap filling exercise.
We provide a general MIP formulation for \ctest generation and specify the optimization problem for a state-of-the-art model. 
A user study with 40 participants across four generation strategies shows that our approach significantly outperforms two baselines and performs on-par with the third.
Our approach further generates \ctests that resonates best with our participants in terms of perceived difficulty.
This could be promising to investigate in future work, as the perceived difficulty can substantially impact a learner's motivation.
Finally, our analysis reveals two further research directions for future work; modeling interdependencies between gaps and making LLMs usable for educational purposes by better controlling their output.


\section{Limitations}
While our proposed approach performed reasonably well and is the only one that provides theoretical guarantees to generate \ctests that fulfill all hard constraints, it suffers from three limitations.

\paragraph{Scaling.} First, it cannot scale indefinitely to larger models as the consumption of computational resources scales exponentially with the increasing model size in the worst case (cf.\,\cref{sec:appendix-mip-primer}).
Moreover, existing solving methods do not transfer well to GPUs due to a limited parallelization.
In comparison, the gap size strategy scales linearly as it is only limited by the number (and length) of the words considered as potential gaps (but does not provide any theoretical guarantees as its solutions are  approximate). 
We explore one possible way to alleviate the scaling issues to some extent by investigating other, mathematically equal formulations of the optimization objective in \cref{sec:appendix-objective} and find that we can substantially reduce the run time.
Another, fundamentally different research direction could be to only include the upper layers of a model in the MIP; considering the output of lower layers as fixed input features (as is done in this work).
This remains to be investigated in future work.

\paragraph{Other architectures.} Second, more complex activation functions such as Gaussian Error Linear Units (GELU) are increasingly being used in recent models, but their respective MIP formulations do not provide any guarantees so far.
This remains an open research question.

\paragraph{GPT-4.} Finally, we used GPT-4 instead of an open source model such as BLOOM~\citep{workshop2023bloom}, considering the accessibility of models for teachers who are interested in using LLMs to generate exercises.
Following this thought, we tried to keep the prompt as simple as possible and tuned it until the model was capable of generating gaps properly (interestingly, asking the model to generate the correct number of blanks resulted in faulty responses).
An evaluation with teachers to gain more insights into how domain experts interact with different \ctest generation systems is ongoing.
This also includes a prompting-specific training with methods such as chain-of-thought prompting~\citep{wei2022chain} and role-play prompting~\citep{kong2023better}.

\section{Ethical Considerations}
\paragraph{Data collection.} The study fulfills all conditions of our university's guidelines for ethical research and has been approved by a spokesperson of the ethics committee of our university.
To ensure a GDPR-conform data collection, we do not collect any personal data of our participants. 
Before participation, every participant is informed about the collected data, its usage, and instructed on how their data can be accessed, edited, deleted post-study.
Participation is only possible upon consent; if not provided, any collected data such as the hash of the study key are immediately deleted.
All our participants were volunteers who participated out of self-interest and received no compensation. 
Although this made the recruitment of participants more difficult, we conjectured that this would result in more motivated participants (and responses of higher quality), in contrast to setups where their main motivation is some form of compensation (money, course credit, or something else).
All data is anonymized for publication.

\paragraph{Risks.} Our findings with respect to GPT-4 show that although the model struggles with fulfilling hard constraints, it can still generate convincing (but misleading) explanations. 
This emphasizes that the use of LLMs in the educational domain requires careful consideration, especially in the context of self-learning where no teacher is present.
Finally, we note that the models we investigate in this research are primarily developed for English.
While we provide a general formulation of the optimization problem in \cref{sec:prelim}, this requires further adaptation to language-specific models which may be difficult to obtain especially for endangered languages. 
However, we note that the considerable performance of the XGB model with a rather small training dataset could provide a chance; easing the adaptation of our approach to other languages.

\section*{Acknowledgments}
We thank Hendrik Schuff for the insightful discussions about the user study setup and Max Glockner for the interesting discussions on the general storyline. 
We further thank Jiahui Geng,  Jan-Christoph Klie, Toru Sasaki, and Martin Tutek for their helpful feedback on the paper draft.
Finally, we thank our anonymous reviewers who provided insightful feedback and engaged in discussions.
This work has been funded by the LOEWE Distinguished Chair ``Ubiquitous Knowledge Processing'', LOEWE initiative, Hesse, Germany (Grant Number: LOEWE/4a//519/05/00.002(0002)/81) and the German Research Foundation (DFG) for the project ``Globally Optimal Neural Network Training'' within the SPP 2298.

\bibliography{custom}
\bibliographystyle{acl_natbib}

\clearpage

\appendix

\section{Introduction}

\subsection{Infeasibility of Brute Force}\label{sec:appendix-brute-forcing}
The large number of possible combinations makes it infeasible to estimate the difficulty of every possible \ctest and to select the one that comes closest to the target difficulty. 
For instance, only considering the gap placement without varying the gap size results in $\binom{n}{m}=\frac{n!}{m! (n-m)!}$ \ctests (for placing $m$ gaps across $n$ words).   
The average run time of the 200 multi-layer perceptrons (MLP) we evaluated during our hyperparameter tuning (\cref{sec:appendix-hyperparameter}) is 1.06 ms for each gap and 22.36 ms for the whole \ctest (using an \textit{Intel Core™ i5-8400 CPU} with 6 x 2.80GHz). 
Consequently, trying out all possible \ctest combinations for 20 out of 40 possible gaps would already require approximately 97.74 years.

\subsection{Primer on Mixed-Integer Programming}\label{sec:appendix-mip-primer}
Similar to integer linear programming (ILP), the goal of mixed-integer programming (MIP) is to identify an optimal solution for a given problem that is described by a mathematical model.\footnote{This section provides a brief introduction into MIP. For more details, we refer the interested reader to \citet{schrijver1986theory} and \citet{wolsey1998integer}.}
The model generally consists of three components~\citep[p.\,2]{nocedal1999numerical}.
First, \textit{variables} that each need to be assigned some value to form a potential solution.
Second, an \textit{objective function} of the variables that needs to be minimized or maximized.
Third, \textit{constraints} that need to be satisfied; represented by (in)equalities and functions containing the variables. 
In contrast to the output generated by LLMs, the global solution of MIPs provides mathematical guarantees on the optimality of the solution and the preservation of the constraints.
In other words, as long as the model is feasible (i.e., contains no contradicting constraints) the found solution can be proven to be optimal and to satisfy all constraints.

\paragraph{Methods.}
Commonly used methods for solving MIP problems with above mathematical guarantees are branch-and-bound~\citep{land1960automatic}, cutting planes~\citep{gomory1960algorithm}, and their combination (branch-and-cut).
These algorithms consider the whole solution space of the problem but in a much more efficient manner than brute force.
They furthermore ensure the optimality of the solution by only taking feasible solutions into account.
To reduce the solution space efficiently, they only search for solutions that are better than the already found ones.
This is done by solving a relaxation at each node that provides an estimate whether in the subproblem corresponding to the node, a better solution can be found. 
If not, the node is cut off; effectively reducing the solution space.
Although these methods are usually much more efficient than brute force, one substantial limitation is the worst-case complexity.

\paragraph{Complexity.}
A key challenge of MIP is that ensuring the mathematical guarantees also results in NP-hard problems.
As such, the general worst-case complexity has been shown to be exponential (see, e.g., \citealt{dash2002exponential}) except for some cases with a proven polynomial worst-case complexity (see, e.g., \citealt{basu2023complexity}). 
Despite the high theoretical worst-case complexity, various methods have been developed to ensure that practical instances are solved in reasonable time.
One such method is presolving~\citep{achterberg2020presolve}.
Presolving transforms the problem into one that is easier to solve, but has the same properties (theoretically) regarding feasibility and boundness; and furthermore, shares the same optimal solution.  
Presolving can substantially reduce the problem complexity, making MIP practically feasible.
To show the practical feasibility of our approach, we provide statistics on the problem size and the impact of presolving, branch-and-bound, and cutting planes in \cref{sec:appendix-empirical-complexity}.
We further discuss the importance of the actual problem formulation and devise alternate formulations to improve the run time of \texttt{MIP} in \cref{sec:appendix-objective}.

\subsection{\texttt{GPT-4}: Prompt and Response (\cref{fig:constraints-example})}\label{sec:appendix-gpt-4-simple}
\begin{figure*}[!htb]
  \begingroup
  \small
  \arrayrulecolor{gray}
  \newcolumntype{L}[1]{>{\small\raggedright\arraybackslash}m{#1}}
  \begin{tabular}{|*{1}{@{ }L{.98\textwidth}@{ }|}}
    \hline
    Generate a C-Test with difficulty 0.1 and 3 gaps using following sentence:
    
    The cat sat on the bench today.
    \\ 
    \hline 
    A C-Test is a type of language proficiency test where words in a given text are partially deleted, and test-takers are asked to complete the missing parts based on the context. The difficulty is often set by the number and choice of gaps.
    
    For a difficulty of 0.1 with 3 gaps from the sentence "The cat sat on the bench today.", the C-test can be constructed as follows:
    
    The c\_\_ sat \_\_ the b\_\_ today.
    
    The solutions are:
    
        c\_\_ = cat \\
        \_\_ = on \\
        b\_\_ = bench \\
    
    Depending on the target audience, the difficulty might vary, but this provides a basic structure for a C-Test from the given sentence.
    \\
    \hline
  \end{tabular}
  \endgroup
  \caption{Prompt (top) and response (bottom) of GPT-4~\citep{openai2023gpt4} for the request to turn a short sentence into a \ctest. As can be seen, the word \textit{on} is fully turned into a gap, showing that the model fails to follow all generation constraints for \ctests.}
  \label{fig:gpt-4-simple-example}
\end{figure*}

\cref{fig:gpt-4-simple-example} shows the prompt and response we used to generate a \ctest out of a short sentence. 
We find that GPT-4 fails to follow all constraints for \ctest generation; more specifically, that only parts of a word should be turned into a gap. 
Interestingly, the model provides a good explanation of the construction principle of \ctests but cannot follow it.

\section{Task Instantiation}

\subsection{Feature description}\label{sec:appendix-feature-description}
\cref{tab:feature-description} shows the full list of the features we used to train our gap error-rate prediction model $f_{\theta}$ (cf. \citealt{Beinborn2016}, pp.~219--224 for a full description of the features).
Note, that the feature \textit{PositionOfGap} is described as ``[t]he position of the gap (indicated as the number of tokens)'' (\citealt{Beinborn2016}, page 224).
In other words, this is the index of the respective token which is a constant that does not depend on the placement of the other gaps.
Originally, \textit{LengthOfSolutionInCharacters} describes the length of the word in characters that would remain constant (\citealt{Beinborn2016}, page 220). 
To provide our model with a notion of the gap size, we change this feature to the length of the gap in characters.
Both BERT-based features use the BERT$_{\text{base}}$ model (marked by $^\ddag$) and were proposed by \citet{lee-etal-2020-empowering}.

\begin{table*}[!htb]
    \centering
    \begin{small}
    \begin{tabular}{llccl}
        \toprule
        Index $k$ & Feature & Size & Placement & Type \\ 
        \midrule
        0 & AvgSentenceLength & - & - & Float \\ 
        1 & AvgWordLengthInCharacters & - & - & Float \\ 
        2 & AvgWordLengthInSyllables & - & - & Float \\ 
        3 & BigramSolutionRank & - & - & Integer \\ 
        4 & COPCognate\_Exists & - & - & Binary \\ 
        5 & GapIsADJ & - & - & Binary \\ 
        6 & GapIsADV & - & - & Binary \\ 
        7 & GapIsART & - & - & Binary \\ 
        8 & GapIsCONJ & - & - & Binary \\ 
        9 & GapIsNN & - & - & Binary \\ 
        10 & GapIsNP & - & - & Binary \\ 
        11 & GapIsPP & - & - & Binary \\ 
        12 & GapIsPR & - & - & Binary \\ 
        13 & GapIsV & - & - & Binary \\ 
        14 & IsAcademicWord & - & - & Binary \\ 
        15 & IsCompound & - & - & Binary \\ 
        16 & IsDerivedAdjective & - & - & Binary \\ 
        17 & IsFunctionWord & - & - & Binary \\ 
        18 & IsInflectedAdjective & - & - & Binary \\ 
        19 & IsInflectedNoun & - & - & Binary \\ 
        20 & IsInflectedVerb & - & - & Binary \\ 
        21 & IsLemma & - & - & Binary \\ 
        22 & IsWordWithLatinRoot & - & - & Binary \\ 
        23 & LanguageModelProbability & - & - & Float \\ 
        24 & LanguageModelProbabilityOfPrefix & - & - & Float \\ 
        25 & LanguageModelProbabilityOfSolution & - & - & Float \\ 
        26 & LeftBigramLogProbability & - & - & Float \\ 
        27 & LeftTrigramLogProbability & - & - & Float \\ 
        28 & LmRankOfSolution & - & - & Float \\ 
        29 & MaxStringSimWithCandidate & - & - & Float \\ 
        30 & NrOfBigramCandidates & - & - & Integer \\ 
        31 & NrOfCandidates & - & - & Integer \\ 
        32 & NrOfTrigramCandidates & - & - & Integer \\ 
        33 & NrOfUbySenses & - & - & Integer \\ 
        34 & NrOfUnigramCandidates & - & - & Integer \\ 
        35 & NumberOfChunksPerSentence & - & - & Float \\ 
        36 & OccursAsText & - & - & Binary \\ 
        37 & PhoneticScore & - & - & Float \\ 
        38 & PhoneticSimilarity & - & - & Float \\ 
        39 & RightBigramLogProbability & - & - & Float \\ 
        40 & RightTrigramLogProbability & - & - & Float \\ 
        41 & TrigramLogProbability & - & - & Float \\ 
        42 & TrigramSolutionRank & - & - & Integer \\ 
        43 & TypeTokenRatio & - & - & Float \\ 
        44 & Uby\_XDiceScore & - & - & Float \\ 
        45 & UnigramLogProbability & - & - & Float \\ 
        46 & UnigramSolutionRank & - & - & Float \\ 
        47 & VerbVariation & - & - & Float \\ 
        48 & posProbability & - & - & Float \\ 
        49 & $^\ddag$BERT$_{\texttt{base-cased}}$ word prediction probability  & \cmark & - & Float \\ 
        50 & $^\ddag$BERT$_{\texttt{base-cased}}$ entropy(softmax(top50))  & \cmark & - & Float \\ 
        51 & NumberOfGapsInCoverSentence & - & \cmark & Integer \\ 
        52 & NumberOfPrecedingGaps & - & \cmark & Integer \\ 
        53 & NumberOfPrecedingGapsInCoverSentence & - & \cmark & Integer \\ 
        54 & OccursAsGap & - & \cmark & Binary \\ 
        55 & PositionOfGap & - & - & Integer \\ 
        56 & IsCompoundBreak & \cmark & - & Binary \\ 
        57 & IsReferentialGap & \cmark & - & Binary \\ 
        58 & IsSyllableBreak & \cmark & - & Binary \\ 
        59 & LengthOfSolutionInCharacters & \cmark & - & Integer \\ 
        60 & LengthOfSolutionInSyllables & - & - & Integer \\ 
        \bottomrule
    \end{tabular}
    \end{small}
    \caption{Features of our model $f_{\theta}$. \cmark marks a dependency of the feature on the size or placement of the gap. For clarity, we use the same nomenclature as \citet{Beinborn2016}.}
    \label{tab:feature-description}
\end{table*}

\section{Gap Difficulty Model}

\begin{figure}
    \centering
    \begin{subfigure}{\columnwidth}
    \caption{Masked-regression.}\label{fig:mlr}
    \includegraphics[width=0.9\textwidth]{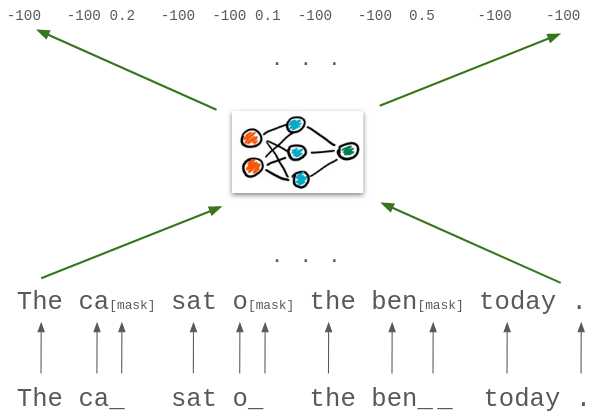}
    \end{subfigure}
    
    \begin{subfigure}{\columnwidth}
    \caption{[CLS] token prediction.}\label{fig:cls}
    \includegraphics[width=0.9\textwidth]{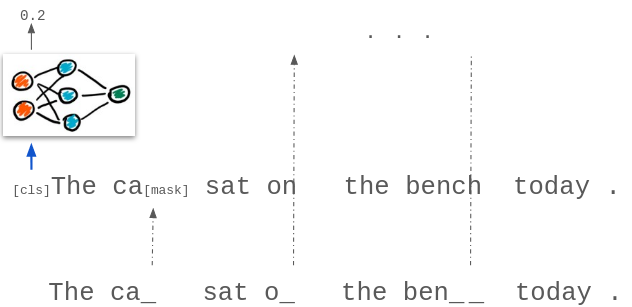}
    \end{subfigure}
    
    \begin{subfigure}{\columnwidth}
    \caption{Feature-enriched [CLS] token prediction.}\label{fig:cls-feature}
    \includegraphics[width=0.9\textwidth]{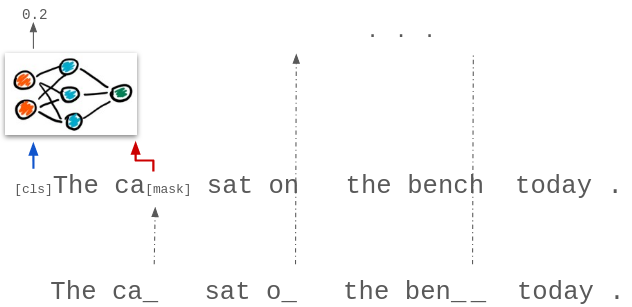}
    \end{subfigure}

    \caption{Transformer-based model setups.}\label{fig:transformers}
\end{figure}

\subsection{Transformer-based Models}\label{sec:appendix-transformers}
We evaluate three different setups for the Transformer-based models (cf. \cref{fig:transformers}).

\begin{description}[noitemsep,topsep=3pt,itemsep=3pt,itemindent=-1em]
\item[MR] Our first setup trains the model in a masked-regression (MR) manner.
For each gap in the \ctest, we insert the \texttt{[mask]} token.
We then process the whole sentence as is; i.e., we tokenize it and feed resulting sequence into the model which is then trained to predict the gap error rate for each masked token.
We further use a special label (-100) with a modified loss function to ensure that only the gaps are considered during training.  
An example is shown in \cref{fig:mlr}.
\item[CLS] Our second setup only uses the \texttt{[cls]} token for gap error rate prediction.
To do so, we turn each of the gap into a sentence where only the gap itself is masked and use the \texttt{[cls]} in a sentence regression manner (cf. \cref{fig:cls}).
\item[CLS+F] Finally, we additionally enrich the CLS setting with the 59 features proposed by \citet{beinborn-2014-ctest}. 
For each gap we then concatenate the \texttt{[cls]} token with the feature vector to predict the gap error rate (cf. \cref{fig:cls-feature}).
Note, that we exclude both BERT-based features in this setup, assuming that the models already have sufficient knowledge about the respective prediction probabilities of the masked gap. 
\end{description}

\subsection{Hyperparameter Tuning}\label{sec:appendix-hyperparameter}
We mainly tune hyperparameters for the SVM and MLP.
For the SVM, we evaluate $c \in $ \{0.00001, 0.0001, 0.001, 0.01, 0.1, 1, 2, 4, 8, 16, 32, 64, 128, 256, 512, 1024, 2048, 10000, 100000\} and find that $c=0.01$ performs best.
For the MLP, we randomly generate 100 configurations where we sample $\{1,2\}$ layers with $\{10,\dots,100\}$ hidden units and run each configuration twice, once with Linear and once with ReLU activation functions; using a batch size of 10.
All neural models are further trained for 250 epochs with an initial learning-rate of 5e-5 using the AdamW optimizer~\citep{loshchilov2019decoupled} and the mean squared error (MSE) loss.
The transformer-based models are trained with a batch size of 5.
We use the development set (cf. \cref{tab:ctest-data-statistics}) to select our model for evaluation on the test set; i.e, we use the model from the best performing epoch.
The best performing MLP configuration for the Linear activation function has two layers with 87 hidden units in the first, and 91 in the second layer.
The best performing MLP configuration for the ReLU activation function has 31 hidden units in the first and 16 in the second layer (with a total of two layers).
We do not tune any hyperparameters for the XGB model, using the default configuration.
We use Scikit-learn (LR and SVM, \citealt{scikit-learn}), the XGBoost library~\citep{chen2016xgboost}, and PyTorch~\citep{paszke2019pytorch} with the transformers library~\citep{wolf-etal-2020-transformers} to implement and train our models.

\subsection{Results}\label{sec:appendix-full-results-intrinsic}

\begin{table*}[!t]
    \centering
    \begin{small}
    \begin{tabular}{lrrrr}
        \toprule
        & \multicolumn{2}{c}{Dev} & \multicolumn{2}{c}{Test}\\ \cmidrule(r){2-3}\cmidrule(l){4-5}
        Model& $\downarrow$ RMSE &  $\uparrow$ Pearson's $\rho$ & $\downarrow$ RMSE & $\uparrow$ Pearson's $\rho$ \\
        \midrule 
        SVM \citep{beinborn-2014-ctest} & \textbf{0.23} & 0.50 & --- & --- \\
        SVM \citep{lee-etal-2019-manipulating} & 0.24 & 0.49 & --- & --- \\
        MLP \citep{lee-etal-2019-manipulating} & 0.25 & 0.42 & --- & ---\\
        BiLSTM \citep{lee-etal-2019-manipulating} & 0.24 & 0.49 & --- & --- \\ 
        \midrule
        SVM$_{\text{Linear}, c=0.01}$ & 0.270 & 0.485 & 1.846 & $-0.111$ \\
        MLP$_{\text{ReLU}}$ & 0.242 & 0.518 & 1.113 & 0.099 \\
        MLP$_{\text{Linear}}$ & \textbf{0.232} & 0.548 & 1.681 & 0.085 \\
        LR & 0.239 & 0.559 & 2.949 & $-0.385$ \\
        XGB & 0.237 & \textbf{0.614} & \textbf{0.285} & \textbf{0.439} \\ 
        \midrule
        BERT$^{\text{MR}}_{\text{base}}$ & 0.249 & 0.420 & 0.311 & 0.279 \\
        BERT$^{\text{MR}}_{\text{large}}$ & 0.255 & 0.362 & 0.319 & 0.174 \\
        RoBERTa$^{\text{MR}}_{\text{base}}$ & 0.258 & 0.388 & 0.324 & 0.159 \\
        RoBERTa$^{\text{MR}}_{\text{large}}$ & 0.262 & 0.281 & 0.324 & 0.050 \\
        DeBERTa$^{\text{MR}}_{\text{v3-base}}$ & 0.248 & 0.502 & 0.311 & 0.245  \\
        DeBERTa$^{\text{MR}}_{\text{v3-large}}$ & 0.245 & 0.484 & 0.308 & 0.259 \\
        \midrule
        BERT$^{\text{CLS}}_{\text{large}}$ & 0.263 & 0.270 & 0.708 & 0.006 \\
        BERT$^{\text{CLS}}_{\text{base}}$ & 0.259 & 0.348 & 0.598 & 0.006 \\
        RoBERTa$^{\text{CLS}}_{\text{base}}$ & 0.268 & 0.287 & 0.456 & -0.031 \\
        RoBERTa$^{\text{CLS}}_{\text{large}}$ & 0.262 & 0.284 & 0.573 & 0.014 \\
        DeBERTa$^{\text{CLS}}_{\text{v3-base}}$ & 0.268 & 0.283 & 0.469 & -0.022  \\
        DeBERTa$^{\text{CLS}}_{\text{v3-large}}$ & 0.267 & 0.277 & 0.491 & 0.016 \\
        \midrule
        BERT$_{^{\text{CLS}+\text{F}}\text{large}}$ & 0.239 & 0.592 & 85.055 & -0.001 \\
        BERT$^{\text{CLS}+\text{F}}_{\text{base}}$ & 0.245 & 0.570 & 60.432 & -0.030 \\
        RoBERTa$^{\text{CLS}+\text{F}}_{\text{base}}$ & 0.239 & 0.587 & 100.689 & -0.004 \\
        RoBERTa$^{\text{CLS}+\text{F}}_{\text{large}}$ & 0.241 & 0.592 & 82.148 & 0.022 \\
        DeBERTa$^{\text{CLS}+\text{F}}_{\text{v3-base}}$ & 0.243 & 0.579 & 72.880 & 0.012  \\
        DeBERTa$^{\text{CLS}+\text{F}}_{\text{v3-large}}$ & 0.241 & 0.585 & 64.031 & -0.015 \\
        \bottomrule
    \end{tabular}
    \end{small}
    \caption{Root mean squared error (RMSE) and Pearson's correlation $\rho$ for predicting the gap error-rate across different models. The first four rows show the results reported in the respective work (hence, there are no results on the test portion of our data); all other rows report the results of our experiments. All results are averaged over ten runs with different random seeds. Overall, we find that XGB performs best on the test data.} 
    \label{tab:results-model-tuning-full}
\end{table*}

\cref{tab:results-model-tuning-full} shows the full results of our gap difficulty model experiments on the development and test data. 
Overall, we find that XGB performs best on the test set in terms of RMSE and Pearson's $\rho$, and has the highest Pearson's $\rho$ on the development set. 
Whereas MLP$_{\text{Linear}}$ has a lower RMSE on the development set, we find that the results substantially worsen on the test set, indicating that the models have overfitted on the training data.
Interestingly, the feature-enriched Transformer-models seem to suffer from the same overfitting as the standard feature-based models, as they perform substantially worse on the test set.
Finally, we find that the MR setup performs surprisingly well, showing the overall best performance across all Transformer-based setups, even outperforming the simpler CLS token setup.
This indicates that considering the gap interdependencies (as done in the MR setup but not in the CLS setup) provides crucial knowledge to the model and should be investigated in more detail in future work.

\subsubsection{Error analysis of DeBERTa$_{\text{v3-large}}$}

\begin{figure*}[!tb]
    \centering
    \include{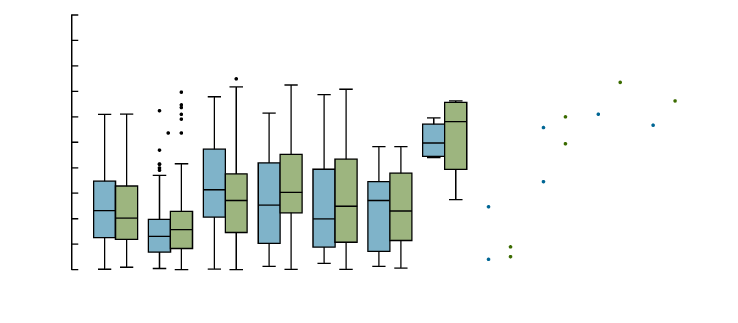}    
    \vspace{-2em}
    \caption{Absolute differences between the predicted to the true error-rate ($\Delta$ gap error-rate) sorted by gap size.}\label{fig:error-analysis-intrinsic}  
\end{figure*}

\begin{figure}[!tb]
    \centering
    \include{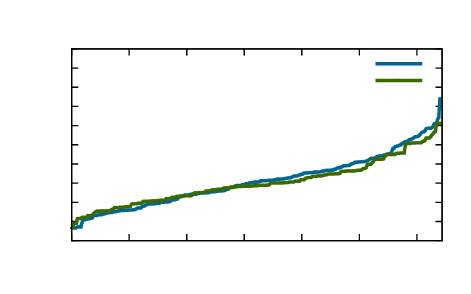}    
    \vspace{-2em}
    \caption{Predicted gap error-rates of XGB and DeBERTa$_{\text{v3-large}}$. We find that XGB covers a larger range of predicted error-rates than DeBERTa$_{\text{v3-large}}$, especially with respect to the maximum predicted values (0.74 vs 0.61).}\label{fig:predicted-error-rates}  
\end{figure}

\cref{fig:error-analysis-intrinsic} shows box plots on the differences between the predicted and true error-rates (absolute values).
We can see that DeBERTa$_{\text{v3-large}}$ performs better than XGB for the gap sizes one and three, but worse for gap sizes equal or larger than four. 
Plotting all predicted error-rates (\cref{fig:predicted-error-rates}) shows that XGB covers a larger range with minimum and maximum values at 0.07 and 0.74 in contrast to DeBERTa$_{\text{v3-large}}$ with minimum and maximum values at 0.07 and 0.61. 
We further find that DeBERTa$_{\text{v3-large}}$ tends to predict too low error-rates for 61.56\% of the gaps whereas for XGB, this happens in only 57.81\% of the cases, indicating that DeBERTa$_{\text{v3-large}}$ tends to underestimate the difficulty of the gaps.

\subsubsection{Feature Ablation for XGB}\label{sec:appendix-ablation}

\begin{table*}[!t]
    \centering
    \begin{small}
    \begin{tabular}{lrrrr}
        \toprule
        & \multicolumn{2}{c}{Dev} & \multicolumn{2}{c}{Test}\\ \cmidrule(r){2-3}\cmidrule(l){4-5} \\
        Feature& $\downarrow$ RMSE & $\uparrow$ Pearson's $\rho$ & $\downarrow$ RMSE & $\uparrow$ Pearson's $\rho$ \\
        \midrule
        All & 0.240 & 0.533 & 0.285 & 0.439 \\ 
        \midrule
        $-$ Readability & \textbf{+0.023} & \textbf{+0.090} &-0.051 & -0.353 \\
        \quad \makecell[l]{$-$ Surface} & -0.006 & -0.003 & -0.044 & -0.290 \\
        \quad \makecell[l]{$-$ Lexical-semantic} & -0.019 & -0.052 & -0.047 & -0.480 \\
        \quad \makecell[l]{$-$ Syntactic} & -0.001 & -0.006 & -0.018 & -0.089 \\
        \midrule 
        $-$ Word Difficulty & -0.020 & -0.154 & -0.068 & -0.285 \\
        \quad \makecell[l]{$-$ Familiarity} & -0.020 & -0.106 & -0.043 &- 0.386 \\
        \quad \makecell[l]{$-$ Morphology \\ \quad \& Compounds} & \textbf{+0.003} & \textbf{+0.056} & -0.019 & -0.089  \\
        \quad \makecell[l]{$-$ Syntax \\ \quad \& Context} & -0.013 & \textbf{+0.013} & -0.030 & -0.241 \\
        \quad \makecell[l]{$-$ L1 Influence} & -0.010 & -0.100 & -0.036 & -0.307 \\
        \quad \makecell[l]{$-$ Spelling Difficulty} & -0.029 & -0.123 & -0.034 & -0.265 \\
        \midrule
        $-$ Candidate Ambiguity & -0.011 & -0.002 & -0.092 & -0.279 \\
        \quad \makecell[l]{$-$ Micro-level} & -0.034 & -0.110 & -0.044 & -0.471 \\
        \quad \makecell[l]{$-$ Macro-level} & -0.010 & -0.027 & -0.104 & -0.230 \\
        \midrule
        $-$ Item Dependency & -	0.005 & \textbf{+0.006} & -0.053 & -0.465 \\
        \quad \makecell[l]{$-$ Position} & \textbf{+0.005} & \textbf{+0.022} &  -0.038 & -0.350 \\
        \quad \makecell[l]{$-$ Neighbor Effects} &-0.004 & \textbf{+0.019} & -0.052 & -0.406 \\
        \quad \makecell[l]{$-$ Referentiality} & \textbf{+0.006} & \textbf{+0.020} &-0.038 & -0.376 \\
        \midrule
        $-$ BERT & -0.015 & -0.096 & -0.052 & -0.349 \\ 
        \bottomrule
    \end{tabular}
    \end{small}
    \caption{Ablation study for different subsets of features using the XGB model. Interestingly, removing some of the features improves the model's performance on the development data but  substantially deteriorates the performance on the test data.}
    \label{tab:results-feature-ablation}
\end{table*}

While the features proposed by \citet{Beinborn2016} are linguistically and pedagogically motivated, they have not been evaluated using XGB as the model.
We thus conduct an ablation study to identify any features that may be removed without affecting the model's performance. 
For ablation, we follow the taxonomy of \citet{Beinborn2016}, who propose four feature categories:
\begin{description}[noitemsep,topsep=3pt,itemsep=3pt,itemindent=-1em]
\item[Readability] Captures the difficulty of the overall text. They are further categorized into surface-level ($k$=0,\,1,\,2), lexical-semantic ($k$=43,\,47), and syntactic ($k$=35) features.
\item[Word Difficulty] Captures the individual difficulty of a word which is comprised of familiarity ($k$=33,\,45,\,59,\,60), morphology \& compound ($k$=15,\,16,\,18--21,\,56,\,58), syntax \& context ($k$=5--13,\,17,\,26,\,39,\,41,\,48), L1 influence ($k$=4,\,14,\,22,\,44), and spelling difficulty ($k$=23--35,\,36--38) features.
\item[Candidate Ambiguity] Considers ambiguity introduced by possible solution candidates and is divided into mirco-level (i.e., close-range; $k$=3,\,29--32,\,34,\,42,\,46) and marco-level (i.e., long-range; $k$=28) features.
\item[Item Dependency] Captures dependencies between different items by using position ($k$=51--53,\,55), neighbor ($k$=27,\,40,\,54), and referentiality ($k$=57) features. 
\end{description}
In addition, we conduct an extra ablation experiment for the BERT-based ($k$=49,\,50) features proposed by \citet{lee-etal-2020-empowering}.
\cref{tab:results-feature-ablation} shows the impact of removing a whole feature category or individual feature subsets on the XGB model.  
Interestingly, we can observe multiple instances where removing a set of features improves the model's performance on the development data; especially the removal of all readability features that improves the rooted mean squared error (RMSE) by 0.023 and Pearson's $\rho$ by 0.09. 
Moreover, their impact on the performance seems to be interdependent, as their individual removal  decreases both scores. 
We can also see that the removal of item dependency features (and subsets) consistently improves Pearson's $\rho$.
In contrast, we observe substantial drops especially in terms of Pearson's $\rho$ on the test set which deviates from the commonly used static \ctest format present in the training and development sets.
We thus conclude that the features proposed by \citet{Beinborn2016} and \citet{lee-etal-2020-empowering} substantially assist the model in generalizing to different gap sizes and placements.

\section{User Study}

\subsection{Setup}\label{sec:appendix-preliminary}
This section provides further details with respect to the setup of our user study.


\begin{figure}[!th]
\begin{small}
    \centering 
    \include{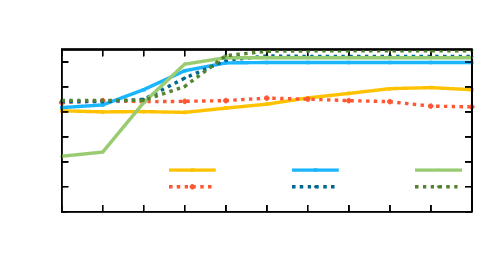}
    \vspace{-2em}
    \label{fig:preliminary-levenshtein}
\end{small}
    \caption{Edit distance (total gap size per \ctest); averaged across 100 \ctests for different target difficulties.}
    \label{fig:preliminary}
\end{figure}

\subsubsection{Reimplementing \texttt{SEL} and \texttt{SIZE}}\label{sec:appendix-reimplementation}
Given the high performance of the XGB model, reimplementing the \texttt{SEL} and \texttt{SIZE} strategies with the XGB model may result in better \ctests.
We thus reimplement both strategies using the XGB model and estimate the performance between different generation strategies by measuring the variability in terms of edit distance of the resulting \ctest (i.e., the total number of characters turned into a gap).

\paragraph{Original baselines.}
\citet{lee-etal-2019-manipulating} define \texttt{SEL} and \texttt{SIZE} as follows:
\begin{description}[noitemsep,topsep=3pt,itemsep=3pt,itemindent=-1em]
\item[\texttt{SEL}] estimates the difficulty of all $n=40$ candidate gaps using an SVM trained on the 59 hand-crafted features defined by \citet{Beinborn2016}. 
All gaps are then divided into two sets; each set only consists of gaps that are easier (or harder) than the target difficulty $\tau$.
Each set is then sorted according to the difference (i.e., the distance) between a gap's difficulty and $\tau$.
Finally, the \ctest is generated by selecting the gaps closest to $\tau$ from each set in an alternating manner, until the final number of gaps $m=20$ have been selected.
In the reimplementation, we replace the original model with our XGB model.
\item[\texttt{SIZE}] estimates the gap difficulty of all default gaps (using the same model as in SEL).
In addition, \citet{lee-etal-2019-manipulating} train two SVMs that predict the relative change in difficulty if the gap was increased (decreased) by one character.
To reduce the time required for feature processing, both models only use six features that least degrade the performance (identified in an ablation study). 
Using these models, the gap size is then either increased (decreased) until the target difficulty $\tau$ is achieved. 
For the reimplementation, we replace the relative prediction model with our XGB model and directly  predict the difficulty of a gap that is increased (or decreased) by one character.
This is made possible due to some improvements (in terms of run time efficiency) we added to the feature pipeline provided by \citet{Beinborn2016}.\footnote{We obtained permission to share an executable \texttt{.jar} file of the improved pipeline under an open source license.}
As in the original algorithm, we do this until we achieve a gap difficulty of larger (smaller) than $\tau$.
\end{description}

\paragraph{Experimental setup.}
Note, that we cannot use our test data as it has been generated by the very same generation strategies (\texttt{SEL} and \texttt{SIZE}) we aim to evaluate and would distort the results.
For the experiments, we randomly sample 100 passages from the GUM corpus~\citep{Zeldes2017}.
As this corpus was designed as an educational corpus for pedagogical use, we conjecture that it better fits our use case than other open corpora comprised of news articles or Wikipedia articles.
We further investigate the helpfulness of varying the BERT features ($k=49$ and $k=50$) as recomputing them leads to the highest overhead in terms of computation time (see \cref{sec:appendix-runtime-mip}).
\paragraph{Results.}
\cref{fig:preliminary} shows the results of our approach with (\texttt{MIP}$_{\text{BERT}}$) and without (\texttt{MIP}$_{\text{no-BERT}}$) varying the BERT features, \texttt{SEL} and \texttt{SIZE} as proposed by \citet{lee-etal-2019-manipulating}, and our reimplementation using the XGB model (\texttt{SEL}$_{\text{XGB}}$ and \texttt{SIZE}$_{\text{XGB}}$).
Overall, we find that varying the BERT features leads to a higher variability with smaller edit distances for lower target difficulties, and larger edit distances for higher target difficulties.
We further find that \texttt{SEL} results in substantially smaller edit distances for lower difficulties than \texttt{SIZE}$_{\text{XGB}}$; indicating that the relative difficulty model proposed by \citet{lee-etal-2019-manipulating} may lead to better \ctests especially for easy ones.
We do not see major differences between \texttt{SEL} and \texttt{SEL}$_{\text{XGB}}$, but find that \texttt{SEL} results in slightly lower minimum edit distances.
For the study, we thus use the \texttt{SEL} and \texttt{SIZE} implementations without any changes, as they display a higher variability, and hence, are stronger baselines.

\subsubsection{Run time of \texttt{MIP}, \texttt{SEL}, and \texttt{SIZE}}\label{sec:appendix-runtime-mip}
While the optimality of the solution found by \texttt{MIP} is ensured by algorithm design (cf.\,\cref{sec:appendix-mip-primer}), one frequent limitation of constraint optimization approaches are their potentially long run times until finding a feasible solution.
To show the feasibility of our approach, we measure the run time of the gap variability experiments conducted in the previous section (\cref{sec:appendix-reimplementation}).
All experiments were conducted using an \textit{Intel Core™ i5-8400 CPU} with 6 x 2.80GHz.
\paragraph{Results.}
For MIP$_{\text{BERT}}$, the solver requires 22.5 seconds on average to find an optimal solution.
We further find that not varying the BERT features can lead to a substantial speedup by 16.9 seconds; reducing the run time of the solver to 5.6 seconds.
In addition, computing the varying BERT features without a GPU leads to an additional overhead of 25.5 seconds.
In total, not varying the BERT features thus reduces the run time by 42.4 seconds due to the smaller number of decision variables ($2 \cdot 20$) and static BERT features.
Nonetheless, we find that the 48.6 seconds of MIP$_{\text{BERT}}$ against the 6.2 seconds of MIP$_{\text{no-BERT}}$ still remains within an acceptable range; especially when considering the better results of MIP$_{\text{BERT}}$.
The run times of \texttt{SEL}  and \texttt{SIZE} are on average 14.6 seconds and 15.3 seconds, respectively; resulting in a difference of $\sim30$ seconds between \texttt{MIP} and the baselines.\footnote{While we cannot provide run time measurements for \texttt{GPT-4} due to too many instabilities regarding the latency, we provide some estimates based on Llama-2~\citep{touvron2023llama}. We find that our prompts would consist of 1277.75 input tokens (on average). Using the model and the inference speed reported by \citet[p.\,48]{touvron2023llama}, then results in an approximate run time of $1277.75$ \text{tokens} $\cdot 25 \frac{\text{ms}}{\text{token}} \approx 31.9$ seconds on eight NVIDIA A100 GPUs with 80 GB RAM.}

\paragraph{Discussion.}
Although the difference of $\sim30$ seconds may seem substantial at first, \texttt{MIP} has two inherent advantages over \texttt{SIZE} (\texttt{SEL}). 
First, the theoretical guarantees that come with the mixed-integer formulation ensure that under a given model $f_\theta$, it is guaranteed that the resulting C-Test is optimal. 
Same cannot be considered for all other baselines, as the solution space is reduced approximatively for \texttt{SIZE} (\texttt{SEL}). 
Moreover, selecting or excluding specific words as well as limiting the gap size can be done by simply adding constraints while the very same guarantees are being kept (i.e., the resulting solution is optimal for model $f_\theta$). 
For instance, turning a word at position $i$ into a gap of size $j$ only requires adding the constraint $s_{i,j}=1$. 
Second, \texttt{MIP} considers the whole solution space during generation. 
Consequently, this allows us to add constraints across the whole solution space. 
In contrast, changing the gap placement in \texttt{SIZE} (gap size in \texttt{SEL}) would require a manual assessment and adaptation of the C-Test as the placement (gap size) is not considered during generation.
Besides the improvements in terms of optimization objective we propose in \cref{sec:appendix-objective}, the run time---especially for the feature generation---could be further improved by incorporating more efficient models such as TinyBERT~\citep{jiao-etal-2020-tinybert} in addition with other methods that improve inference efficiency~\citep{treviso-2023-efficiency}. 
This remains to be investigated in future work.

\subsubsection{Empirical Complexity of MIP}\label{sec:appendix-empirical-complexity}

\begin{table}[!t]
    \centering
    \begin{small}
    \begin{tabular}{lrr@{$\pm$}r}
        \toprule
         & $\mu$ & & $\sigma$ \\
        \midrule
        \multicolumn{4}{l}{Model Statistics} \\ 
        \midrule
        \quad \makecell[l]{Rows} & 4,413.95 & & 4.02 \\ 
        \quad \makecell[l]{Columns} & 179,491.45 && 24.06 \\ 
        \quad \makecell[l]{Nonzeros} & 179,486.50 && 566.61 \\ 
        \midrule
        Constraints & 21,887.62 & & 1,484.86 \\ 
        \midrule
        \multicolumn{4}{l}{Variables (before presolve)} \\ 
        \midrule
        \quad \makecell[l]{Continuous} & 6,563.00 && 0.00 \\ 
        \quad \makecell[l]{Integer} & 172,928.45 && 24.06 \\ 
        \quad \makecell[l]{Binary} & 172,928.45 && 24.06 \\ 
        \midrule 
        \midrule
        \multicolumn{4}{l}{Presolved} \\ 
        \midrule 
        \quad \makecell[l]{Rows} & 8,854.53 && 749.26 \\ 
        \quad \makecell[l]{Columns} & 6,852.54 && 550.05 \\ 
        \quad \makecell[l]{Nonzeros} & 46,695.43 && 5,788.68 \\ 
        \midrule
        \multicolumn{4}{l}{Variables (after presolve)} \\ 
        \midrule
        \quad \makecell[l]{Continuous} & 1,248.73 && 101.67 \\ 
        \quad \makecell[l]{Integer} & 5,603.81 && 459.85 \\ 
        \quad \makecell[l]{Binary} & 5,496.23 & & 455.90 \\ 
        \midrule
        \midrule
        Cutting Planes & 4,823.34 && 3,219.46 \\ 
        \midrule
        B\,\&\,B Nodes & 2,336.07 && 5,916.47 \\ 
        \bottomrule
    \end{tabular}
    \end{small}
    \caption{Average ($\mu$) statistics about the optimization model, number of constraints, and variables. Presolving consistently and substantially reduces the number of variables. Most affected by the individual problem complexity are the the number of visited nodes and cutting planes with a high standard deviation ($\sigma$).}  
    \label{tab:emirical-complexity}
\end{table}

As discussed in \cref{sec:appendix-mip-primer}, the employed methods do not affect the worst-case complexity and the problem itself remains NP-hard.  
However, analyzing the actual complexity of the experiments conducted in \cref{sec:appendix-reimplementation} reveals that practical instances are being solved in reasonable time.
As can be seen in \cref{tab:emirical-complexity}, the standard deviations with respect to the model statistics, the number of constraints, and variables are low; indicating a stable level of problem complexity.
Moreover, we find that presolving consistently and substantially reduces the problem complexity. 
Finally, we can observe that the individual problem complexity primarily affects the final steps of solving the problem; i.e., the number of cutting plane cuts and visited branch-and-bound (B\,\&\,B) nodes.
In practice, all our experiments terminated and in feasible time (cf.\,\cref{sec:appendix-objective}).

\subsubsection{Text Selection}\label{sec:appendix-reading-difficulties}

\begin{table}[!tb]
    \centering
    \begin{small}
    \begin{tabular}{lrrrrc}
    \toprule
    $\mathcal{T}$ & Metric & $\mu$ & $\sigma$ & \ctest & \\
    \midrule
        $\mathcal{T}_1$ & ARI & 21.04 & 8.49 & 34.21 & \xmark \\ 
        & Coleman-Liau & 12.65 & 1.87 & 12.84 & \ymark \\ 
        & DaleChallIndex & 11.26 & 0.98 & 11.97 & \ymark \\ 
        & FleschReadingEase & 38.58 & 23.02 & 10.37 & \xmark \\ 
        & GunningFogIndex & 22.24 & 6.95 & 33.08 & \xmark \\ 
        & Kincaid & 17.28 & 6.97 & 27.88 & \xmark \\ 
        & LIX & 64.52 & 17.59 & 93.02 & \xmark \\ 
        & RIX & 10.35 & 4.90 & 18.33 & \xmark \\ 
        & SMOGIndex & 16.96 & 3.69 & 21.71 & \xmark \\ 
        \midrule
        $\mathcal{T}_2$ & ARI & 10.3 & 3.85 & 14.12 & \ymark \\ 
        & Coleman-Liau & 9.43 & 2.21 & 10.63 & \ymark \\ 
        & DaleChallIndex & 9.64 & 0.94 & 9.54 & \ymark \\ 
        & FleschReadingEase & 69.26 & 15.77 & 57.86 & \ymark \\ 
        & GunningFogIndex & 12.87 & 3.32 & 16.43 & \xmark \\ 
        & Kincaid & 8.69 & 3.39 & 11.90 & \ymark \\ 
        & LIX & 41.93 & 8.91 & 51.44 & \xmark \\ 
        & RIX & 4.42 & 1.90 & 6.60 & \xmark \\ 
        & SMOGIndex & 11.27 & 2.19 & 13.68 & \xmark \\ 
        \midrule
        $\mathcal{T}_3$ & ARI & 14.96 & 6.14 & 21.07 & \ymark \\ 
        & Coleman-Liau & 10.33 & 1.77 & 8.67 & \ymark \\ 
        & DaleChallIndex & 10.12 & 0.80 & 9.30 & \xmark \\ 
        & FleschReadingEase & 57.85 & 16.36 & 54.60 & \ymark \\ 
        & GunningFogIndex & 17.16 & 4.99 & 21.70 & \ymark \\ 
        & Kincaid & 12.4 & 4.96 & 16.76 & \ymark \\ 
        & LIX & 51.52 & 12.31 & 62.63 & \ymark \\ 
        & RIX & 6.52 & 2.84 & 8.00 & \ymark \\ 
        & SMOGIndex & 13.71 & 2.67 & 14.29 & \ymark \\ 
        \midrule
        $\mathcal{T}_4$ & ARI & 14.98 & 5.19 & 15.92 & \ymark \\ 
        & Coleman-Liau & 11.14 & 1.51 & 10.45 & \ymark \\ 
        & DaleChallIndex & 10.24 & 0.77 & 9.87 & \ymark \\ 
        & FleschReadingEase & 53.85 & 14.27 & 55.65 & \ymark \\ 
        & GunningFogIndex & 17.33 & 4.27 & 17.91 & \ymark \\ 
        & Kincaid & 12.65 & 4.22 & 13.23 & \ymark \\ 
        & LIX & 52.52 & 10.99 & 55.22 & \ymark \\ 
        & RIX & 6.94 & 2.83 & 7.50 & \ymark \\ 
        & SMOGIndex & 14.19 & 2.37 & 14.29 & \ymark \\ 
    \bottomrule
    \end{tabular}
    \end{small}
    \caption{Readability scores of $\mathcal{T}_1$--$\mathcal{T}_4$ on eight different metrics.}
    \label{tab:reading-difficulties}
\end{table}

As we expect most of our participants to have at least a college degree, we focus on books that presumable have a higher reading difficulty.
The selected books are (in alphabetical order):
\begin{itemize}
    \item[$\mathcal{T}_1$] Beyond Good and Evil (Friedrich Nietzsche) 
    \item[$\mathcal{T}_2$] Crime and Punishment (Fyodor Dostoevsky)
    \item[$\mathcal{T}_3$] Emma (Jane Austen)
    \item[$\mathcal{T}_4$] Pride and Prejudice (Jane Austen)
\end{itemize}
To evaluate how well the randomly sampled passages in the study represent each book, we conduct an analysis with respect to their reading difficulty and compare each passage against all possible passages that follow our selection criteria (cf. \textbf{Text selection} in \cref{sec:study}).
\cref{tab:reading-difficulties} shows the scores for eight reading difficulty metrics that were computed using a respective python package.\footnote{\url{https://pypi.org/project/readability/}}
For each text ($\mathcal{T}$), we show the average ($\mu$) and standard deviation ($\sigma$) for the reading difficulty of all paragraphs and the one used in the study (\ctest). 
The last column indicates if the paragraph used in the study falls within (\ymark) or outside (\xmark) the boundary of $\mu\pm\sigma$.
Overall, we see that all selected text passages have a high reading difficulty satisfying our criteria of college graduate level or higher reading difficulty.
Moreover, we see that the text passages (except $\mathcal{T}_1$) fall within the boundary of average and standard deviation across a majority of metrics, indicating that they represent the text well in terms of reading difficulty.

\subsection{Design}\label{sec:appendix-design}

\subsubsection{Group Configurations}\label{sec:appendix-hypercube}

\cref{tab:study-design} shows the eight configurations used in our study where each participant solves \ctests generated from each model $\mathcal{M}$, text $\mathcal{T}$, and two target difficulties $\tau$.

\begin{table}[!htb]
    \centering
    \begin{small}
    \begin{tabular}{lllllllll}
        \toprule
        & \multicolumn{2}{c}{$\mathcal{T}_1$} & \multicolumn{2}{c}{$\mathcal{T}_2$} & \multicolumn{2}{c}{$\mathcal{T}_3$} &\multicolumn{2}{c}{$\mathcal{T}_4$}  \\
        & $\text{M}$ & $\tau$ & $\text{M}$ & $\tau$ & $\text{M}$ & $\tau$ & $\text{M}$ & $\tau$ \\
        \midrule 
        1 & $\text{M}_1$ & 0.1  & $\text{M}_4$ & 0.9  & $\text{M}_2$ & 0.1  & $\text{M}_3$ & 0.9  \\
        2 & $\text{M}_1$ & 0.9  & $\text{M}_2$ & 0.1  & $\text{M}_4$ & 0.9  & $\text{M}_3$ & 0.1  \\     
        3 & $\text{M}_2$ & 0.1  & $\text{M}_1$ & 0.9  & $\text{M}_3$ & 0.9  & $\text{M}_4$ & 0.1  \\     
        4 & $\text{M}_2$ &  0.9  & $\text{M}_1$ & 0.1  & $\text{M}_3$ & 0.1  & $\text{M}_4$ & 0.9  \\  
        5 & $\text{M}_3$ & 0.1  & $\text{M}_2$ & 0.9  & $\text{M}_4$ & 0.1  & $\text{M}_1$ & 0.9  \\   
        6 & $\text{M}_3$ & 0.9  & $\text{M}_4$ & 0.1  & $\text{M}_1$ & 0.9  & $\text{M}_2$ & 0.1  \\
        7 & $\text{M}_4$ & 0.1  & $\text{M}_3$ & 0.9  & $\text{M}_2$ & 0.9 & $\text{M}_1$ & 0.1  \\        
        8 & $\text{M}_4$ & 0.9  & $\text{M}_3$ & 0.1  & $\text{M}_1$ & 0.1  & $\text{M}_2$ & 0.9  \\
        \bottomrule
    \end{tabular}
    \end{small}
    \caption{Configurations used in our user study. As can be seen, all generation strategies $\text{M}$ and target difficulties $\tau$ are evenly distributed. Our models are: \texttt{GPT-4} ($\text{M}_1$), \texttt{MIP} ($\text{M}_2$), \texttt{SEL} ($\text{M}_3$), and \texttt{SIZE} ($\text{M}_4$).} 
    \label{tab:study-design}
\end{table}

\subsubsection{Study Examples and Questionnaire}\label{sec:appendix-study-examples}
\begin{figure*}[!htb]
    \centering
    \includegraphics[width=0.8\textwidth]{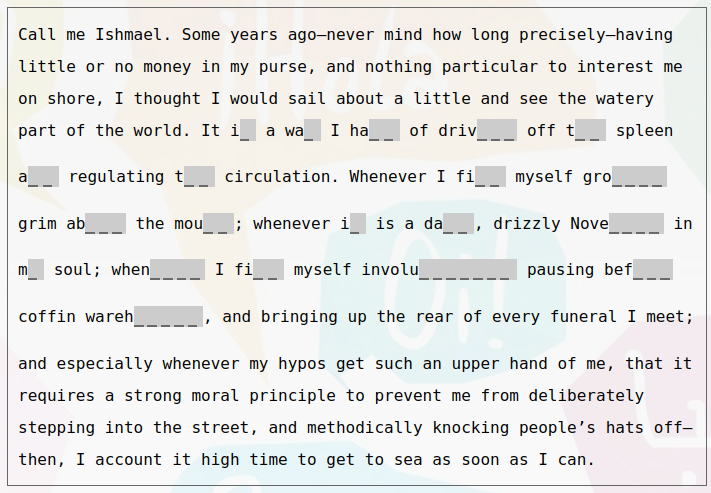}
    \caption{Example \ctest showing the interface used in our user study.}
    \label{fig:study-ctest}
\end{figure*}

\begin{figure*}[!htb]
    \centering
    \includegraphics[width=0.8\textwidth]{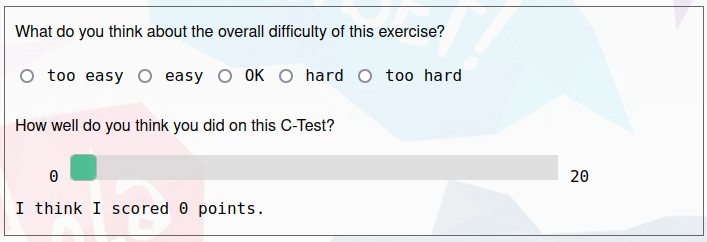}
    \caption{Collected feedback after each \ctest.}
    \label{fig:study-feedback}
\end{figure*}

\cref{fig:questionnaire} shows all questions participants were asked to answer upon registration. 
An example \ctest is provided in \cref{fig:study-ctest}.
\cref{fig:study-feedback} shows the feedback our participants were asked to provide after each \ctest.

\begin{figure}[!htb]
\newcommand{\BOX}[1]{\parbox[b]{#1}{\hrulefill}}
\fbox{\parbox{\linewidth}{\raggedright
\textbf{Q1}: Please estimate your current language proficiency in English \\
\textbf{A1}: $\Circle$~\textit{Beginner (A1)} $\Circle$~\textit{Elementary (A2)} $\Circle$~\textit{Intermediate (B1)} $\Circle$~\textit{Upper Intermediate (B2)} $\Circle$~\textit{Advanced (C1)} $\Circle$~\textit{Proficient (C2)} \medskip\\ 
\textbf{Q2}: I studied English for about \BOX{.5cm} years. \medskip\\
\textbf{Q3}: How often do you practice or speak English? \\
\textbf{A3}: $\Circle$ \textit{Never} $\Circle$ \textit{Monthly} $\Circle$ \textit{Weekly} $\Circle$ \textit{Daily} \medskip\\
\textbf{Q4}: What is your native language?  \\
\textbf{A4}: \BOX{4cm}\medskip\\
\textbf{Q5}: Have you tried learning languages (other than English)? If yes, than which ones?  \\
\textbf{A5}: \textit{$\Circle$ Yes, \BOX{1.5cm}.}  $\Circle$ \textit{No.} 
}}
\caption{Study questionnaire.}\label{fig:questionnaire}
\end{figure}

\subsection{Results}\label{sec:appendix-detailed-results}
This section provides a detailed analysis about our study participants and their responses in the questionnaire, as well as details about the GAMM formulation and the resulting p-values.

\subsubsection{Participants}\label{sec:appendix-study-participants}
\begin{figure*}[!htb]
    \begin{minipage}{0.45\textwidth}
    \centering
    \include{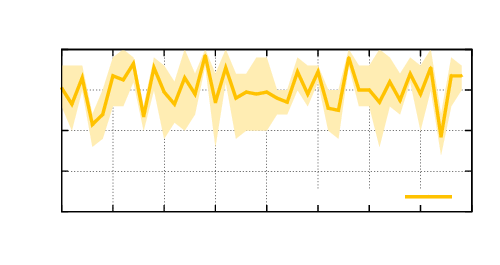}    
    \vspace{-3em}
    \caption{Average score for each participant sorted by their CEFR self-estimate. Shaded areas indicate the maximum and minimum scores.}\label{fig:user-scores}
    \end{minipage}
    \hfill 
    \begin{minipage}{0.45\textwidth}
    \centering
    \include{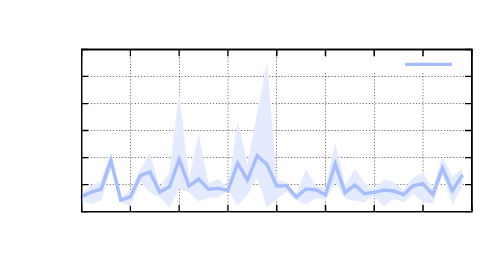}
    \vspace{-3em}
    \caption{Average time taken for each participant sorted by their CEFR self-estimate. Shaded areas indicate the maximum and minimum time taken.}\label{fig:user-times}
    \end{minipage}   
\end{figure*}

The study was distributed across our university and took place between the 21st of August 2023 and 8th of September 2023.
All our participants are volunteers (and received no compensation) with an academic background with at least one college degree.
They have achieved at least B1 proficiency in English based on the common European framework of references for languages (CEFR, \citealt{cefr}).
Overall, two participants stated to have a B1 proficiency, nine a B2 proficiency, and 16 and 13 a C1 and C2 proficiency, respectively. 
On average, our participants have used English for 10 ($\pm 6.5$) years; mostly on a daily (27) or weekly (8) basis.
Only two participants use English on a monthly basis, while three responded with never.
Most of our participants are native German speakers (26).
Other native languages were Chinese (3), Russian (2), Turkish (2), Arabic, Croatian, Italian, Hindi, Korean, Kyrgyz (together with Russian), Spanish, and Vietnamese (1 each). 
Finally, the majority (36) of our participants have attempted to learn a different language---on average 1.97 languages excluding English.
\cref{fig:user-scores} and \cref{fig:user-times} show the average scores and time taken for each participant, sorted by their provided CEFR self-estimate.
Interestingly, we do not find any significant differences between the proficiency, score, and time taken, indicating that our participants actually had a similar English proficiency.
This is in line with our observations from the GAMM analysis that indicates that none of our smooth terms play a significant role for modeling (cf.~\cref{sec:appendix-study-gamm}).

\begin{figure*}[!htb]
    \begin{minipage}{0.45\textwidth}
    \centering
    \begin{tikzpicture}

  \begin{axis}[
    ybar,
    xtick distance = 1,
    symbolic x coords = {B1, B2, C1, C2},
    nodes near coords,
    bar width = 27pt,
  ]\addplot[jgreen!20!black,fill=jgreen!80!white] coordinates {
	(B1,2) 
	(B2,9) 
	(C1,16) 
	(C2,13)
};
\end{axis}

\end{tikzpicture}
    \vspace{-2em}
    \caption{CEFR self-estimates of our participants.}\label{fig:user-cefr}
    \end{minipage}
    \hfill 
    \begin{minipage}{0.45\textwidth}
    \centering
    \begin{tikzpicture}

  \begin{axis}[
    ybar,
    xtick distance = 1,
    symbolic x coords = {never, monthly, weekly, daily},
    nodes near coords,
    bar width = 27pt,
  ]\addplot[ablue!20!black,fill=ablue!80!white] coordinates {
	(never,3) 
	(monthly,2) 
	(weekly,8) 
	(daily,27)
};
\end{axis}

\end{tikzpicture}
    \vspace{-2em}
    \caption{Frequency our participants use English.}\label{fig:user-frequency}
    \end{minipage}   
\end{figure*}

\begin{figure*}[!htb]
    \centering
    \begin{small}
    \begin{tikzpicture}
  \begin{axis}[
    width=1\textwidth,
    height=0.3\textheight,
    ybar,
    xmin=-0.5,
    xmax=28.5,
    xtick distance = 1,
    nodes near coords,
    bar width = 8pt,
  ]\addplot[cyelllow!20!black,fill=cyelllow!80!white] coordinates {
    (0,4) 
    (4,1) 
    (5,1) 
    (6,1)
    (7,3) 
    (8,6) 
    (9,4) 
    (10,8)
    (11,2) 
    (14,1) 
    (15,2) 
    (18,1)
    (19,1) 
    (20,2) 
    (22,1) 
    (25,1)
    (28,1) 
};
\end{axis}
\end{tikzpicture}
    \end{small}
    \vspace{-2em}
    \caption{Number of years our participants have been using English.}\label{fig:user-years}  
\end{figure*}
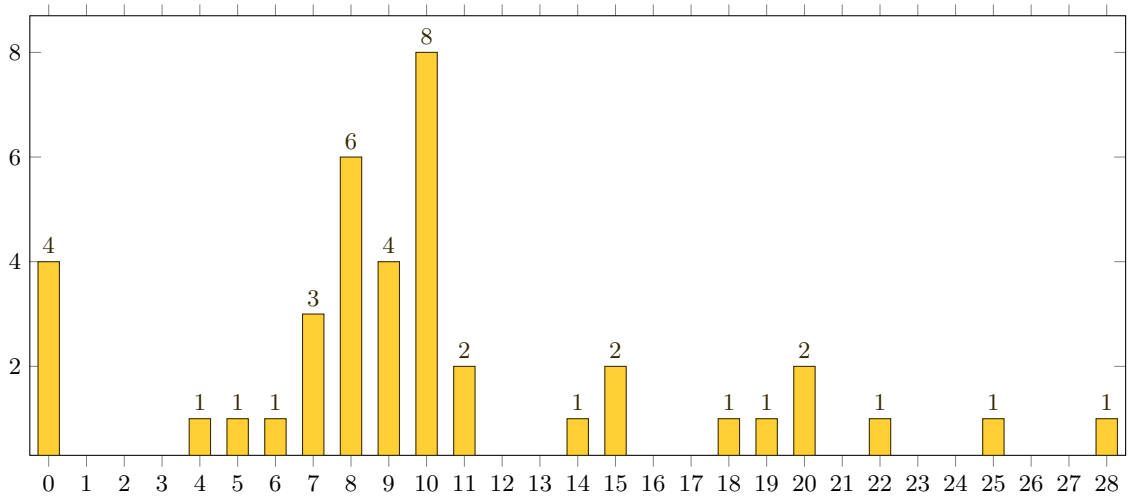

\subsubsection{GAMM Details for $\hat{\tau}$ }\label{sec:appendix-study-gamm}
We formulate our GAM model for the actual difficulty $\hat{\tau}$ as follows:
\begin{align}\label{eq:gam}
\begin{split}
    \hat{\tau} = & \text{M} + \tau + f_1(\mathcal{T})\cdot \text{Z}_1 + f_2(\text{CEFR})\cdot \text{Z}_2  \\
                & \quad + f_3(\text{User})\cdot \text{Z}_3 + f_4(\text{Years}) \\
                & \quad  + \text{M} \cdot \tau + \epsilon,
\end{split}
\end{align}
where M represents the generation strategy and $\tau$ the target difficulty.
$\mathcal{T}$, CEFR, User, and Years are terms for categorical values of text, language proficiency, participant, and number of years participants have been using English, modeled as smooth terms $f(\cdot)$ with a random effect Z.\footnote{Note, that we do not require a random effect for the years as they are quantifiable numeric values.} 
Finally, $\text{M} \cdot \tau$ models the interaction between generation strategy and target difficulty and $\epsilon$ is an unknown vector of random errors.

We use the \texttt{mgcv}~\citep{wood2016smoothing} implementation available in R~\citep{r2013r}.
\cref{fig:gamm-significance} show the p-values between different generation strategies computed using the Wald test~\citep{wald1943tests}. 
Analyzing our GAMM shows that the model's explanatory power is substantial with $R^2 = 0.59$. 
\cref{tab:study-gamm-fixed} shows the F-test statistics~\citep{bruning1987computational} of the GAM model with respect to the parametric fixed terms.
We find that all fixed terms are statistically significant.
Besides M which has been discussed in the main paper, we further find that $\tau$ has a significant and positive effect on the response variable $\hat{\tau}$ with $\beta = 0.14, 95\% \text{CI} [0.05, 0.23]$ and $ p = 0.002$.
Interestingly, we find that all modeled smooth terms are not significant as shown in \cref{tab:study-gamm-smooth}.
This however, is in line with our findings that indicate that our participants' CEFR self-estimate and their number of years using English do not substantially impact the score.
Finally, the underlying text as well as the participant (User) do not substantially impact the score and thus, can be excluded as confounding factors in our observations.\footnote{Note, that this only applies to this specific study; in future studies, all these factors may impact the outcomes.}

\begin{table}[!tb]
\centering
\begin{tabular}{lrrr}
  \toprule
$\hat{\tau}$ & df & F & p-value \\ 
  \midrule
  M & 3.00 & 19.87 & 1.53e-10$^*$ \\ 
  $\tau$ & 1.00 & 9.57 & 0.00246$^*$ \\ 
  $\text{M} \cdot \tau$ & 3.00 & 9.96 & 6.47e-06$^*$ \\ 
   \bottomrule
\end{tabular}
\caption{F-test statistics for the parametric fixed terms of the GAMM. df denotes the degrees of freedom. All terms are statistically significant.} 
\label{tab:study-gamm-fixed}
\end{table}

\begin{table}[!tb]
\centering
\begin{small}
    \begin{tabular}{lrrrr}
  \toprule
$\hat{\tau}$ & edf & Ref.df & F & p-value \\ 
  \midrule
    $f_1(\mathcal{T})$ & 1.7013 & 4.00 & 0.983 & 0.990 \\
    $f_2(\text{CEFR})$ & 0.8445 & 1.00 & 131.858 & 0.997 \\
    $f_2(\text{User})$ & 27.3925 & 40.00 & 4.396 & 0.998 \\
    $f_2(\text{Years})$ & 1.0000 & 1.00 & 0.013 & 0.908 \\
   \bottomrule
\end{tabular}
\end{small}
\caption{F-test statistics for the smooth terms of the GAM model. With edf denoting the effective degrees of freedom, and Ref.df the reference degrees of freedom. None of the terms are statistically significant, indicating that they do not have any impact on $\hat{\tau}$.} 
\label{tab:study-gamm-smooth}
\end{table}

\begin{figure}
    \centering
    \begin{tikzpicture}[scale=1.3]
  \foreach \y [count=\n] in {
      {0,15,30,10},
      {15,0,40,70},
      {30,40,0,10},
      {10,70,10,0},
    } {
      \foreach \x [count=\m] in \y {
        \node[fill=ablue!\x!jgreen, minimum size=13mm, text=white] at (\m,-\n) {};
      }
    }

  \foreach \y [count=\n] in {
      {-,3.5e-11,5.5e-5,6.7e-13},
      {3.5e-11,-,0.006,0.562},
      {5.5e-5,0.006,-,9.3e-04},
      {6.7e-13,0.562,9.3e-04,-},
    } {
      \foreach \x [count=\m] in \y {
        \node[fill=none, minimum size=8mm, text=white] at (\m,-\n) {\x};
      }
    }
  \foreach \a [count=\i] in {\texttt{GPT-4},\texttt{MIP},\texttt{SEL},\texttt{SIZE}} {
    \node[minimum size=6mm] at (\i,0) {\a};
  }
  \foreach \a [count=\i] in {\texttt{GPT-4},\texttt{MIP},\texttt{SEL},\texttt{SIZE}} {
    \node[minimum size=6mm] at (0,-\i) {\a};
  } 
  \node[shading = axis, shading angle = 90, top color=ablue, bottom color=jgreen, middle color=ablue!70!jgreen, minimum width=3mm, minimum height=50mm, rectangle] at (4.7,-2.5) {};
  
  \node[fill=none, minimum size=3mm, text=black] at (5.1,-4.4) {0.0};
  \node[fill=none, minimum size=3mm, text=black] at (5.1,-0.5) {1.0};

\end{tikzpicture}
    \vspace{-2em}
    \caption{P-values of the Wald test between different \ctest generation strategies for $\hat{\tau}$. As can be seen, all strategies significantly outperform \texttt{GPT-4}, and \texttt{MIP} significantly outperforms \texttt{SEL}. We find no significant differences between \texttt{MIP} and \texttt{SIZE}.}
    \label{fig:gamm-significance}
    \definecolor{ablue}{HTML}{006795}
\definecolor{jgreen}{HTML}{98CC70}

\begin{tikzpicture}[scale=1.3]
  \foreach \y [count=\n] in {
      {0,5,20,12},
      {5,0,30,70},
      {20,30,0,40},
      {12,70,40,0},
    } {
      \foreach \x [count=\m] in \y {
        \node[fill=ablue!\x!jgreen, minimum size=13mm, text=white] at (\m,-\n) {};
      }
    }

  \foreach \y [count=\n] in {
      {-,2.5e-14,1.7e-05,2.3e-12},
      {2.5e-14,-,3.2e-04,0.540},
      {1.7e-05,3.2e-04,-,0.003},
      {2.3e-12,0.540,0.003,-},
    } {
      \foreach \x [count=\m] in \y {
        \node[fill=none, minimum size=8mm, text=white] at (\m,-\n) {\x};
      }
    }
  \foreach \a [count=\i] in {\texttt{GPT-4},\texttt{MIP},\texttt{SEL},\texttt{SIZE}} {
    \node[minimum size=6mm] at (\i,0) {\a};
  }
  \foreach \a [count=\i] in {\texttt{GPT-4},\texttt{MIP},\texttt{SEL},\texttt{SIZE}} {
    \node[minimum size=6mm] at (0,-\i) {\a};
  } 
  \node[shading = axis, shading angle = 90, top color=ablue, bottom color=jgreen, middle color=ablue!70!jgreen, minimum width=3mm, minimum height=50mm, rectangle] at (4.7,-2.5) {};
  
  \node[fill=none, minimum size=3mm, text=black] at (5.1,-4.4) {0.0};
  \node[fill=none, minimum size=3mm, text=black] at (5.1,-0.5) {1.0};

\end{tikzpicture}
    \vspace{-2em}
    \caption{P-values of the Wald test between different \ctest generation strategies for the perceived difficulty. As can be seen, all strategies significantly outperform \texttt{GPT-4}, and \texttt{MIP} significantly outperforms \texttt{SEL}. We find no significant differences between \texttt{MIP} and \texttt{SIZE}.}
    \label{fig:gamm-significance-feedback}
\end{figure}

\begin{figure*}
    \centering
    \includegraphics[width=0.85\textwidth]{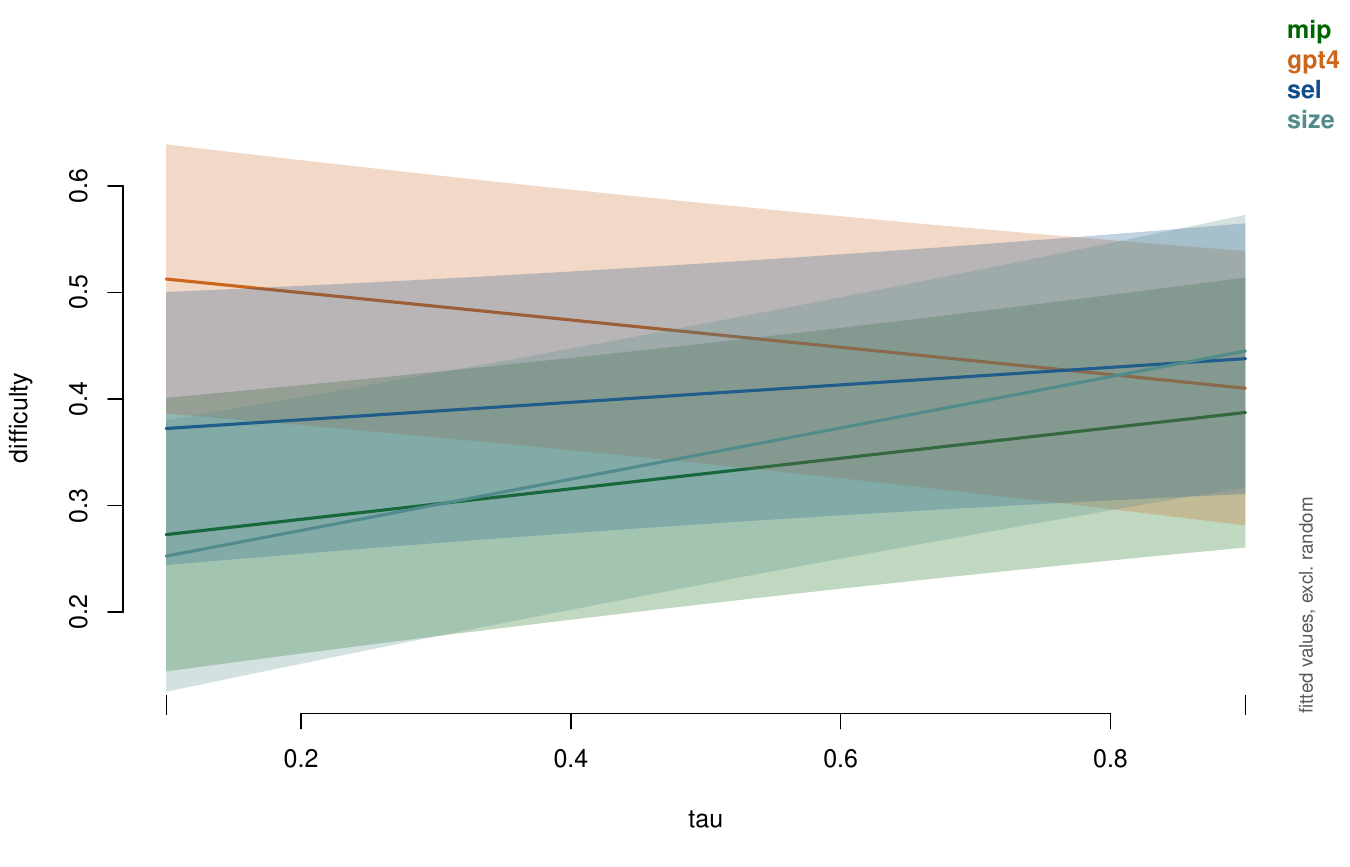}
    \caption{Linear regression curves for each generation strategy and difficulty. As can be seen, \texttt{GPT-4} is the only strategy that shows a negative slope; indicating that the model as an inverted notion of difficulty.}
    \label{fig:difficulty-tendency}
\end{figure*}

\subsubsection{Perceived Difficulty}\label{sec:appendix-perceived-difficulty}
\begin{figure*}[!tb]
    \begin{minipage}{0.45\textwidth}
    \centering
    \include{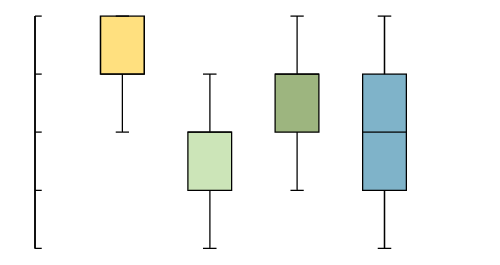}
    \vspace{-2em}
    \caption{Perceived difficulty for $\tau=0.1$ ($\sim$very easy).}\label{fig:user-easy}
    \end{minipage}
    \hfill 
    \begin{minipage}{0.45\textwidth}
    \centering
    \include{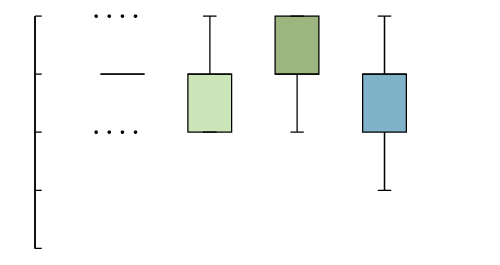}
    \vspace{-2em}
    \caption{Perceived difficulty for $\tau=0.9$ ($\sim$very hard).}\label{fig:user-hard}
    \end{minipage}   
\end{figure*}

\cref{fig:user-easy} and \cref{fig:user-hard} show the box plots of our participant's responses for the perceived difficulty.
We can see that \texttt{MIP} generated \ctests that were perceived easiest by our participants.
In addition, we again find a clear tendency that the easy \ctests generated by \texttt{GPT-4} were also perceived as more difficult than the hard \ctests.
Finally, conducting Wald tests~\citep{wald1943tests} using our GAMM shows that \texttt{MIP} significantly outperform \texttt{SEL} and \texttt{GPT-4} and performs on-par with \texttt{SIZE} (cf.\,\cref{fig:gamm-significance-feedback}).
Note, that \texttt{GPT-4} again performs significantly worse than all other generation strategies.

\paragraph{GAMM statistics.}\label{sec:appendix-study-gamm-feedback}
\cref{tab:study-gamm-fixed-feedback} and \cref{tab:study-gamm-smooth-feedback} show the analysis of our GAM model with respect to the perceived difficulty. 
We observe similar trends as for $\hat{\tau}$ with an even higher substantial explanatory power $\text{R}^2=0.81$.

\begin{table}[!htb]
\centering
\begin{tabular}{lrrr}
  \toprule
Feedback & df & F & p-value \\ 
  \midrule
  M & 3.00 & 21.698 & 2.16e-11$^*$ \\ 
  $\tau$ & 1.00 & 23.871 & 3.07e-06$^*$ \\ 
  $\text{M} \cdot \tau$ & 3.00 & 9.269 & 1.39e-05$^*$ \\ 
   \bottomrule
\end{tabular}
\caption{F-test statistics for the parametric fixed terms of the GAMM. df denotes the degrees of freedom. All terms are statistically significant.} 
\label{tab:study-gamm-fixed-feedback}
\end{table}

\begin{table}[!tb]
\centering
\begin{small}
    \begin{tabular}{lrrrr}
  \toprule
Feedback & edf & Ref.df & F & p-value \\ 
  \midrule
    $f_1(\mathcal{T})$ & 2.825e+00 & 4.00 & 39.469 & 0.843 \\
    $f_2(\text{CEFR})$ & 1.817e-06 & 4.00 & 0.000 & 1.000 \\
    $f_2(\text{User})$ & 2.062e+01 & 40.00 & 1.313 & 0.991 \\
    $f_2(\text{Years})$ & 2.771e+00 & 3.05 & 0.920 & 0.457 \\
   \bottomrule
\end{tabular}
\end{small}
\caption{F-test statistics for the smooth terms of the GAM model, with edf denoting the effective degrees of freedom, and Ref.df the reference degrees of freedom. None of the terms are statistically significant, indicating that they do not have any impact on the perceived difficulty.} 
\label{tab:study-gamm-smooth-feedback}
\end{table}

\subsection{Error Analysis}\label{sec:appendix-error-analysis}

\paragraph{Details \texttt{MIP} vs \texttt{SIZE}.}\label{sec:appendix-mip-vs-size}
\cref{tab:mip-vs-size-chains} show the number of occurrences of successive gaps (i.e., multiple gaps that occur in succession). 
Although it seems reasonable that successive gaps should be more difficult to fill out as less words provide context, we do not see substantial differences between \ctests generated with $\tau=0.1$ and $\tau=0.9$. 
This indicates that the XGB model may not have a good notion about how successive gaps impact the overall \ctest difficulty.

\begin{table*}[!t]
    \centering
    \begin{tabular}{rrrrr}
    \multicolumn{5}{c}{$\tau=0.1$}\\
    \toprule
       \# Succ.  & $\mathcal{T}_1$& $\mathcal{T}_2$& $\mathcal{T}_3$ & $\mathcal{T}_4$ \\ 
        \midrule
       1 & 3 & 5 & 10 & 6 \\ 
       2 & 5 & 1 & 1 & 1 \\ 
       3 & 1 & - & 1 & 1 \\ 
       4 & 1 & - & - & 1 \\ 
       5 & - & - & 1 & 1 \\ 
       6 & - & 1 & - & - \\ 
       7 & - & 1 & - & - \\ 
       8 & - & - & - & - \\ 
       9 & - & - & - & - \\ 
       10 & - & - & - & - \\ 
    \bottomrule
    \end{tabular}
    \hspace{7em}
    \begin{tabular}{rrrrr}
    \multicolumn{5}{c}{$\tau=0.9$}\\
    \toprule
       \# Succ.  & $\mathcal{T}_1$& $\mathcal{T}_2$& $\mathcal{T}_3$ & $\mathcal{T}_4$ \\ 
        \midrule
       1 & 4 & 6 & 7 & 3 \\ 
       2 & 4 & 2 & 2 & 2 \\ 
       3 & - & 2 & - & 1 \\ 
       4 & 2 & 1 & 1 & - \\ 
       5 & - & - & 1 & - \\ 
       6 & - & - & - & - \\ 
       7 & - & - & - & - \\ 
       8 & - & - & - & - \\ 
       9 & - & - & - & - \\ 
       10 & - & - & - & 1 \\ 
         \bottomrule
    \end{tabular}
    \caption{Number of occurrences for successive gaps (\# Succ.) in \ctests generated by \texttt{MIP} for $\tau=0.1$ (left) and $\tau=0.9$ (right).}\label{tab:mip-vs-size-chains}
\end{table*}

\begin{figure*}
  \begingroup
  \small
  \arrayrulecolor{gray}
  \newcolumntype{L}[1]{>{\ttfamily\small\raggedright\arraybackslash}m{#1}}
  \begin{tabular}{|*{2}{@{ }L{.49\textwidth}@{ }|}}
    \hline
    \dots From t\_\_{$\color{jgreen}\blacksquare$} expense of the child, how\_\_\_\_{$\color{jgreen}\blacksquare$}, he was so\_\_{$\color{jgreen!80!red}\blacksquare$} relieved. The bo\_{$\color{jgreen}\blacksquare$} had, wit\_{$\color{jgreen}\blacksquare$} the add\_\_\_\_\_\_\_{$\color{jgreen!90!red}\blacksquare$} softening cl\_\_\_{$\color{jgreen!50!red}\blacksquare$} of a lin\_\_\_\_\_\_{$\color{jgreen!50!red}\blacksquare$} illness of his mo\_\_\_\_{$\color{jgreen!90!red}\blacksquare$}'s, been the me\_\_\_{$\color{jgreen!50!red}\blacksquare$} of a so\_\_{$\color{jgreen!60!red}\blacksquare$} of reconciliations; and Mr. and Mrs. Ch\_\_\_\_\_\_\_{$\color{jgreen!90!red}\blacksquare$}, having no ch\_\_\_\_\_\_{$\color{jgreen!90!red}\blacksquare$} of their ow\_{$\color{jgreen!90!red}\blacksquare$}, nor any ot\_\_\_{$\color{jgreen!90!red}\blacksquare$} young cr\_\_\_\_\_\_{$\color{jgreen!50!red}\blacksquare$} of equal ki\_\_\_\_\_{$\color{red}\blacksquare$} to care for, offered to ta\_\_{$\color{jgreen!90!red}\blacksquare$} the whole ch\_\_\_\_{$\color{jgreen!50!red}\blacksquare$} of the little Fr\_\_\_{$\color{jgreen!60!red}\blacksquare$}
 soon after\dots
    &
    \dots From t\_\_{$\color{jgreen}\blacksquare$} expense o\_{$\color{jgreen}\blacksquare$} the ch\_\_\_{$\color{jgreen}\blacksquare$}, how\_\_\_\_{$\color{jgreen}\blacksquare$}, he w\_\_{$\color{jgreen}\blacksquare$} soon rel\_\_\_\_\_{$\color{jgreen!90!red}\blacksquare$}. The bo\_{$\color{jgreen}\blacksquare$} had, with t\_\_{$\color{jgreen}\blacksquare$} addit\_\_\_\_\_{$\color{jgreen}\blacksquare$} soften\_\_\_{$\color{jgreen!90!red}\blacksquare$} claim o\_{$\color{jgreen}\blacksquare$} a ling\_\_\_\_\_{$\color{jgreen!90!red}\blacksquare$} illness o\_{$\color{jgreen}\blacksquare$} his mo\_\_\_\_{$\color{jgreen}\blacksquare$}'s, be\_\_{$\color{jgreen!60!red}\blacksquare$} the me\_\_\_{$\color{jgreen!80!red}\blacksquare$} of a so\_\_{$\color{jgreen!80!red}\blacksquare$} of recon\_\_\_\_\_\_\_\_\_{$\color{jgreen!80!red}\blacksquare$}; and M\_{$\color{jgreen}\blacksquare$}. and M\_\_{$\color{jgreen}\blacksquare$}. Churchill, having no children of their own, nor any other young creature of equal kindred to care for, offered to take the whole charge of the little Frank soon after\dots
    \\ \hline
    \multicolumn{1}{c}{(a) C-test of $\mathcal{T}_3$ generated with \texttt{GPT-4} for $\tau = 0.1$} & 
    \multicolumn{1}{c}{(b) C-test of $\mathcal{T}_3$ generated with \texttt{GPT-4} for $\tau = 0.9$}
    \\
  \end{tabular}
  \endgroup
  \caption{\texttt{GPT-4} generated \ctests. Colored squares indicate the gap error-rates (0.0 {$\color{jgreen}\blacksquare$}
{$\color{jgreen!90!red}\blacksquare$}
{$\color{jgreen!80!red}\blacksquare$}
{$\color{jgreen!60!red}\blacksquare$}
{$\color{jgreen!50!red}\blacksquare$}
{$\color{red}\blacksquare$} 1.0)}
\label{fig:gpt-4-errors}
\end{figure*}

\paragraph{\texttt{GPT-4} shortcomings.}
\cref{fig:gpt-4-errors} shows the \ctests generated by \texttt{GPT-4} for $\mathcal{T}_3$ with $\tau=0.1$ and $\tau=0.9$.
As the lower gap error-rates indicate, the more difficult \ctest is comprised of easier gaps, including five gaps of gap size one (compared to two for the easier \ctest).
We can further observe that the gaps are primarily clustered around the beginning of the \ctest.
In addition, we find that GPT-4 generates misleading explanation about its notion of difficulty, for instance:  \\

\noindent
\textit{Based on the provided examples and the desired difficulty, we'll try to generate a C-Test with 20 gaps and a difficulty of 0.1. This means that about 10\% of the gaps should be moderately challenging to fill in.}

Although the explanation is convincing, considering the fact that the model simply tries to add (remove) gaps to increase (decrease) the difficulty of the resulting \ctest makes it clear that the models does not have a notion of gap-level difficulty.
However, the provided explanation may even mislead students to believe that they are actually solving a \ctest of a specific difficulty, resulting in a wrong self-assessment.

\paragraph{Gap size vs gap placement.}
Motivated by the substantial differences between \texttt{SEL} and \texttt{SIZE}, we conduct an analysis to see if their differences can be attributed to either one; gap size or gap placement.
To better assess the differences between different strategies, we use the static generation strategy (\texttt{STAT}) as a common denominator to compare different \ctests.
\cref{tab:ctest-gapsize-default} shows the differences in terms of gap size (i.e., the number of characters) compared to \texttt{STAT}. 
Interestingly, we find that \texttt{SIZE} is most similar to \texttt{STAT} whereas \texttt{MIP} is most dissimilar to \texttt{STAT}.
\cref{tab:ctest-placement-default} furthermore shows that \texttt{MIP} has the highest overlap (consistently around 50\%) to all other strategies in terms of gap placement.
Considering that \texttt{SIZE} and \texttt{MIP} show no significant differences but significantly outperform \texttt{SEL} and \texttt{GPT-4}, we conclude that we cannot attribute the differences in performance to solely a difference gap size or placement.
Moreover, our analysis suggests that \ctests with different gap sizes and placements can be equally good, highlighting the importance of considering interdependencies between gaps. 

\begin{table}[!htb]
    \centering
    \begin{small}
    \begin{tabular}{lrrrr}
    \toprule
         &  \texttt{GPT-4} & \texttt{MIP} & \texttt{SEL} & \texttt{SIZE} \\ 
        \midrule
        $\Delta$ & 77 & 96 & 0 & 62 \\ 
        $\mu$ & 9.6 & 12.0 & 0.0 & 7.8 \\ 
        \bottomrule
    \end{tabular}
    \end{small}
    \caption{Gap size differences compared to the static generation strategy. We show the total number of differences ($\Delta$) across all eight \ctests as well as the average difference per \ctest ($\mu$).}\label{tab:ctest-gapsize-default} %
\end{table}

\begin{table}[!htb]
    \centering
    \begin{small}
    \begin{tabular}{lrrrrr}
    \toprule
         &  \texttt{GPT-4} & \texttt{MIP} & \texttt{SEL} & \texttt{STAT} & \texttt{SIZE} \\ 
        \midrule
        \texttt{GPT-4} & 100.0 & - & - & - & - \\ 
        \texttt{MIP} & 48.1 & 100.0 & - & - & - \\ 
        \texttt{SEL} & 43.1 & 51.3 & 100.0 & - & - \\ 
        \texttt{STAT} & 40.6 & 49.4 & 45.6 & 100.0 & - \\ 
        \texttt{SIZE} & 40.6 & 49.4 & 45.6 & 100.0 & 100.0 \\ 
        \bottomrule
    \end{tabular}
    \end{small}
    \caption{Overlap of the gap placements when compared to the static (\texttt{STAT}) generation strategy (in \%). \texttt{STAT} and \texttt{SIZE} use the same gap placements.}\label{tab:ctest-placement-default} 
\end{table}

\subsection{Importance of Problem Formulation}\label{sec:appendix-objective}
As discussed in \cref{sec:appendix-mip-primer}, the actual problem formulation can make a substantial difference in terms of run time.
To investigate if the run time of \texttt{MIP} can be further improved, we evaluate two alternative formulations of our optimization objective ($|\tau - \hat{\tau}|$).
We especially focus on the absolute value function $|\cdot|$, as this can be done in multiple ways of which some are feasible and others are not.\footnote{
For instance, $\sqrt{(\tau - \hat{\tau})^2}$ equally describes $|\tau - \hat{\tau}|$ but introduces a quadratic term, resulting in an infeasible model.}
We evaluate three different optimization objectives in this work:
\begin{description}[noitemsep,topsep=3pt,itemsep=3pt,itemindent=-1em]
\item[Min/Max] Using the minimum and maximum functions: $|\tau - \hat{\tau}| = \max\{\tau, \hat{\tau}\} - \min\{\tau, \hat{\tau}\}$.
\item[Indicator] Using indicator constraints (i.e., if-else constructs):
\[
|\tau - \hat{\tau}| = 
\begin{cases}
    \tau - \hat{\tau}, & \text{if } \tau > \hat{\tau}, \\
    \hat{\tau} - \tau , & \text{otherwise}.
\end{cases} 
\]
\item[PWL] Using piecewise linear functions: \\ $|\tau - \hat{\tau}| = a(\cdot) + b(\cdot) + c(\cdot) + d(\cdot)$, with
    \begin{enumerate}
        \item[] $a(x) = \tau $ \quad ~~,  $ \forall x \leq 0$
        \item[] $b(x) = \tau - \hat{\tau} $, $ \forall 0 < x \leq \tau$
        \item[] $c(x) = \hat{\tau} - \tau $, $ \forall \tau < x \leq 1$
        \item[] $d(x) = 1 - \tau $, $ \forall 1 < x$.
    \end{enumerate}
\end{description}
For evaluation, we repeat the \texttt{MIP}$_\text{BERT}$ run time experiments from \cref{sec:appendix-reimplementation} using 6 Cores of an \textit{AMD EPYC™ 7742 processor} with 2.25GHz each.
To reduce noise from other components in the code (e.g., feature extraction or data loading) we only measure the time required for solving the optimization objective.
\cref{tab:runtime-objective-functions} shows the average time ($\mu$) and standard deviation ($\sigma$) as well as the minimum ($\min$) and maximum ($\max$) run times for each formulation.
Interestingly, we find that the optimization objective has a higher impact on the actual run time than the difference in hardware.
For instance, the Min/Max formulation that was used throughout this work has a run time of $\sim$32 seconds; requiring only 10.5 seconds longer on the \textit{AMD} hardware compared to using an \textit{Intel Core™ i5-8400 CPU} with 6 x 2.80GHz (22.5 seconds; cf.\,\cref{sec:appendix-runtime-mip}).
In contrast, using a different optimization objective results in substantially different run times, namely $\sim$3 seconds for the indicator objective and $\sim$73 seconds for PWL.
The differences become especially large for the worst-case ($\max$) run times, as the Min/Max formulation requires $\sim$11 minutes and PWL even up to 3 hours and 12 minutes.
In contrast, the worst-case run time of the indicator constraint remains low at only $\sim$7 seconds. 
Overall, the indicator formulation substantially closes the gap of \texttt{MIP} to \texttt{SIZE} and \texttt{SEL} (with a total run time of $\sim$15 seconds) and will be published along with the code to alleviate future research.
\looseness=-1

\begin{table}[]
    \centering
    \begin{tabular}{lrrrr}
    \toprule
    Method & $\mu$ & $\sigma$ & $\max$ & $\min$ \\
    \midrule
    Min/Max & 32.02 & 61.52 & 656.12 & 1.33 \\
    Indicator & \textbf{3.12} & \textbf{1.34} & \textbf{7.17} & \textbf{0.95} \\
    PWL & 72.91 & 370.60 &  11,481.52 & 1.15 \\
    \bottomrule
    \end{tabular}
    \caption{Run time of \texttt{MIP}$_\text{BERT}$ using different formulations of our optimization objective (in seconds). We show the average ($\mu$), standard deviation ($\sigma$), maximum ($\max$), and minimum ($\min$) run times.}
    \label{tab:runtime-objective-functions}
\end{table}

\subsection{GPT-4 Example}\label{sec:appendix-gpt-example}
\cref{fig:gpt-4-prompt} shows the prompt we use to construct \ctests using GPT-4. 
For the few-shot examples, we select the instances from the ACL-2019 dataset with the highest (0.655) and lowest (0.09) difficulty and randomly sample one instance per text (\citealt{lee-etal-2019-manipulating} use four texts in total).
\cref{fig:gpt-4-response} shows the respective response we received after five tries. 
We regenerated the response if the model produced a \ctest with less than 20 gaps.
If the model generated more than 20 gaps, we selected the first 20 gaps.
All prompts and responses are provided in the published data.

\begin{figure*}[!htb]
  \begingroup
  \small
  \arrayrulecolor{gray}
  \newcolumntype{L}[1]{>{\small\raggedright\arraybackslash}m{#1}}
  \begin{tabular}{|*{1}{@{ }L{.98\textwidth}@{ }|}}
    \hline

C-Tests are gap-filling exercises where each only the latter part of a word is made into a gap.
A C-Test is generated by placing gaps in an input text.
The difficulty of a C-Test is the percentage of errors a student makes across all gaps.
Each C-Test consists of 20 gaps. 
Gaps are indicated by '\_' .
Here are some examples:

Example 1:

Input Text 1:

The Serge Prokofieff whom we knew in the United States of America was gay, witty, mercurial, full of pranks and bonheur -- and very capable as a professional musician. These qualities endeared him to both the musicians and the social-economic haute monde which supported the concert world of the post-World War 1, era. 

C-Test 1:

The Se\_ Prokofieff wh\_ we kn\_ in the United States of America was g\_ , wi\_ , merc\_ , full of pra\_ a\_ bon\_ -- and very cap\_ as a profes\_ musi\_ . Th\_ qual\_ ende\_ h\_ to bo\_ the musi\_ and the social-economic ha\_ mo\_ which supported the concert world of the post-World War 1 , era .

Difficulty 1: 0.655

Example 2:

Input Text 2:

It is being fought, moreover, in fairly close correspondence with the predictions of the soothsayers of the think factories. They predicted escalation, and escalation is what we are getting. The biggest nuclear device the United States has exploded measured some 15 megatons, although our B-52s are said to be carrying two 20-megaton bombs apiece.

C-Test 2:

It i\_ being fough\_ , moreover , i\_ fairly clos\_ correspondence wit\_ the prediction\_ of t\_ soothsayers o\_ the thin\_ factories . The\_ predicted escalatio\_ , and escalatio\_ is wha\_ we ar\_ getting . T\_ biggest nuclea\_ device t\_ United State\_ has explode\_ measured som\_ 15 megatons , although our B-52s are said to be carrying two 20-megaton bombs apiece . 

Difficulty 2: 0.09

Example 3:

Input Text 3:

Here was a man with an enormous gift for living as well as thinking. To both persons and ideas he brought the same delighted interest, the same open-minded relish for what was unique in each, the same discriminating sensibility and quicksilver intelligence, the same gallantry of judgment.

C-Test 3:

Here w\_ a man w\_ an e\_ gift f\_ living a\_ well a\_ thinking . T\_ both per\_ and id\_ he bro\_ the s\_ delighted inte\_ , the s\_ open-minded relish f\_ what w\_ unique i\_ each , t\_ same d\_ sensibility a\_ quicksilver i\_ , the same gallantry of judgment . 

Difficulty 3: 0.43

Example 4:

Input Text 4:

The Serge Prokofieff whom we knew in the United States of America was gay, witty, mercurial, full of pranks and bonheur -- and very capable as a professional musician. These qualities endeared him to both the musicians and the social-economic haute monde which supported the concert world of the post-World War 1, era. 

C-Test 4:

The Se\_ Prokofieff wh\_ we kn\_ in t\_ United Sta\_ of Ame\_ was g\_ , witty , merc\_ , full o\_ pranks a\_ bonheur -- a\_ very cap\_ as a profes\_ musician . Th\_ qualities ende\_ him t\_ both t\_ musicians a\_ the social-economic ha\_ monde wh\_ supported the concert world of the post-World War 1 , era .

Difficulty 4: 0.3425

Example 5:

Input Text 5:

St. Louis sits in the center of a relatively slow-growing and in some places stagnant mid-continent region. Slackened regional demand for St. Louis goods and services reflects the region's relative lack of purchasing power. Not all St. Louis industries, of course, have a market area confined to the immediate neighborhood.

C-Test 5:

St. Louis si\_ in t\_ center o\_ a relatively slow-growing a\_ in so\_ places stag\_ mid-continent region . Slac\_ regional dem\_ for St. Lo\_ goods a\_ services refl\_ the reg\_ 's relative la\_ of purch\_ power . N\_ all St. Lo\_ industries , o\_ course , ha\_ a market ar\_ confined t\_ the immediate neighborhood . 

Difficulty 5: 0.295
\\
\hline
Now generate a C-Test with 20 gaps and a difficulty of <$\tau$>

Input Text:
\\
\hline
<Text containing all words $w$>
 \\
    \hline
  \end{tabular}
  \endgroup
  \caption{The prompt used for \ctest generation using GPT-4. We first explain the concept of a \ctest and provide five examples selected from the data collected by \citet{lee-etal-2019-manipulating}. We then add an instruction to generate a \ctest of our target difficulty $\tau$ and finally, provide the text passage that contains all words $w$. All \ctests used in the study are generated using the same five few-shot examples.}
  \label{fig:gpt-4-prompt}
\end{figure*}

\begin{figure*}[!htb]
  \begingroup
  \small
  \arrayrulecolor{gray}
  \newcolumntype{L}[1]{>{\small\raggedright\arraybackslash}m{#1}}
  \begin{tabular}{|*{1}{@{ }L{.98\textwidth}@{ }|}}
    \hline
Now generate a C-Test with 20 gaps and a difficulty of 0.1 

Input Text:
\\
\hline
As little as the act of birth comes into consideration in the whole process and procedure of heredity, just as little is "being-conscious" opposed to the instinctive in any decisive sense; the greater part of the conscious thinking of a philosopher is secretly influenced by his instincts, and forced into definite channels.
\\
\hline
C-Test:

As litt\_ as the a\_ of birth com\_ into consider\_ in the w\_ process a\_ procedure o\_ heredity , jus\_ as litt\_ is " being - conscious " opp\_ to the inst\_ in any dec\_ sense ; the gre\_ part o\_ the cons\_ thinking of a philo\_ is sec\_ influenced by h\_ instincts , and for\_ into def\_ channels.

Difficulty: 0.1

(Note: This C-Test is generated with a target difficulty of 0.1, but the actual difficulty experienced by a test taker might differ based on their familiarity with the content and context.)
\\
    \hline
  \end{tabular}
  \endgroup
  \caption{A \ctest generated by GPT-4 for $\mathcal{T}_1$ and $\tau=0.1$ (last row). Note, that this is the fifth regneration, as previous responses of the model contained less than 20 gaps.}
  \label{fig:gpt-4-response}
\end{figure*}

\end{document}